\documentclass[10pt,twocolumn,letterpaper]{article}

\usepackage{cvpr}
\usepackage{overpic}
\usepackage{booktabs}
\usepackage{multirow}
\usepackage{algorithm}
\usepackage{algpseudocode}
\newcommand{\todo}[1]{{\color{red}#1}}

\makeatletter
\newcommand*\bigcdot{\mathpalette\bigcdot@{.5}}
\newcommand*\bigcdot@[2]{\mathbin{\vcenter{\hbox{\scalebox{#2}{$\m@th#1\bullet$}}}}}
\makeatother

\definecolor{cvprblue}{rgb}{0.21,0.49,0.74}
\usepackage[pagebackref,breaklinks,colorlinks,allcolors=cvprblue]{hyperref}

\title{Gaussian Billboards:\\Expressive 2D Gaussian Splatting with Textures}

\author{Sebastian Weiss \space\space Derek Bradley\\
\small{\textnormal{DisneyResearch$\vert$Studios}}
}

\begin{document}

\twocolumn[{
\maketitle
\begin{center}
\centering
\includegraphics[width=\textwidth]{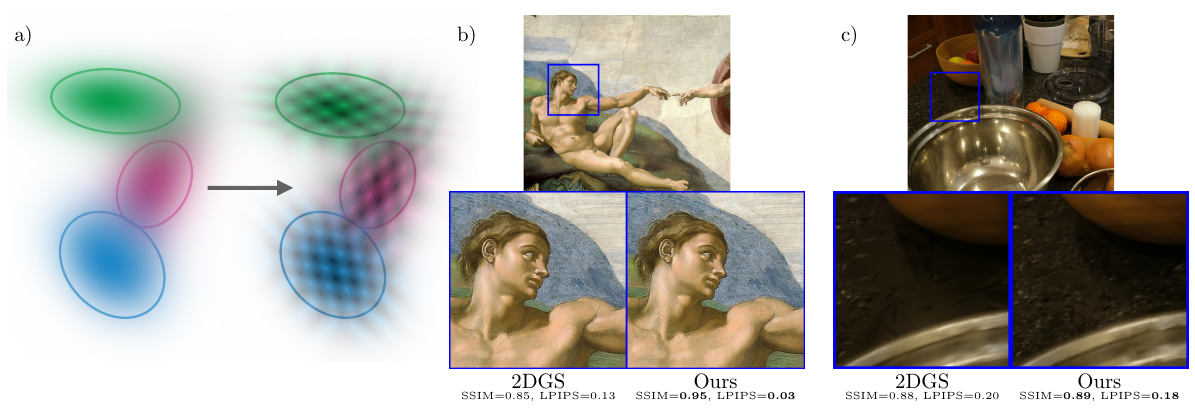}
\captionof{figure}{We propose an enhancement for 2D Gaussian Splatting that (a) replaces the per-primitive solid color with a small per-primitive texture. This allows the 2DGS to represent more details (b) when the number of primitives is fixed to, \eg, 100000. Even with the standard densification strategy and identical hyperparameters, it improves the quality of 3D scene reconstruction (c), shown on a test view.}
\label{fig:teaser}
\end{center}
}]

\newcommand{\figref}[1]{Fig.~\ref{#1}}
\newcommand{\tabref}[1]{Table~\ref{#1}}
\newcommand{\eqnref}[1]{Eq.~\ref{#1}}
\newcommand{\secref}[1]{Section~\ref{#1}}
\newcommand{\appref}[1]{Appendix~\ref{#1}}


\newcommand{\pluseq}{\mathrel{+}=}
\newcommand{\asteq}{\mathrel{*}=}

\newcommand{\shortcite}[1]{\cite{#1}}


\definecolor{algchangedcolor}{RGB}{0,130,0}

\definecolor{dbcolor}{RGB}{50,10,210}
\newcommand\db[1] {{\textcolor{dbcolor}{\em\textbf{DB}: #1}}}

\definecolor{swcolor}{RGB}{10,150,100}
\newcommand\sw[1] {{\textcolor{swcolor}{\em\textbf{SW}: #1}}}

\definecolor{delcolor}{RGB}{210,0,0}
\definecolor{addcolor}{RGB}{0,0,0}
\newcommand\del[1] {{\textcolor{delcolor}{#1}}}
\newcommand\add[1] {{\textcolor{addcolor}{#1}}}

\renewcommand\todo[1] {{\textcolor{red}{\em\textbf{TODO}: #1}}}
\newcommand\camready[1] {{\textcolor{blue}{#1}}}

\makeatletter
\newcommand\footnoteref[1]{\protected@xdef\@thefnmark{\ref{#1}}\@footnotemark}
\makeatother

\renewcommand{\thefootnote}{\fnsymbol{footnote}}

\clubpenalty=10000
\widowpenalty=10000
\displaywidowpenalty=10000

\begin{abstract}\begin{minipage}{\linewidth}
        Gaussian Splatting has recently emerged as the go-to representation for reconstructing and rendering 3D scenes. The transition from 3D to 2D Gaussian primitives has further improved multi-view consistency and surface reconstruction accuracy.
        In this work we highlight the similarity between 2D Gaussian Splatting (2DGS) and billboards from traditional computer graphics. Both use flat semi-transparent 2D geometry that is positioned, oriented and scaled in 3D space. However 2DGS uses a solid color per splat and an opacity modulated by a Gaussian distribution, where billboards are more expressive, modulating the color with a $uv$-parameterized texture.
        We propose to unify these concepts by presenting {\normalfont Gaussian Billboards}, a modification of 2DGS to add spatially-varying color achieved using per-splat texture interpolation.  The result is a mixture of the two representations, which benefits from both the robust scene optimization power of 2DGS and the expressiveness of texture mapping.
        We show that our method can improve the sharpness and quality of the scene representation in a wide range of qualitative and quantitative evaluations compared to the original 2DGS implementation.
    \end{minipage}\end{abstract}
\section{Introduction}
\label{sec:intro}

Neural scene representations have become a standard solution to represent arbitrary 3D scenes from a collection of 2D images, allowing for novel view synthesis, relighting, animation playback or animation retargeting, just to name a few applications.
Recently, 3D Gaussian Splatting (3DGS)~\cite{Kerbl2023-cv} has established itself as the method of choice, outperforming previous representations like NeRF~\cite{Mildenhall2020-mp} or Mixture of Volumetric Primitives~\cite{Lombardi2021-ah}.
3DGS's core idea is to represent the scene explicitly as a collection of volumetric primitives, positioned and oriented in 3D, and then rasterized and rendered as 2D splats on the screen. To make this representation differentiable, the opacity of this volume is modeled via a 3D Gaussian distribution.

Since the introduction of 3DGS, many improvements to its rendering and representation quality have been presented. In particular, 2D Gaussian Splatting (2DGS)~\cite{Huang2024-2dgs} replaces the volumetric primitives with oriented 2D primitives (disks).
Still, representing highly detailed textured areas with potentially low geometric details, \ie an image on a wall, remains challenging.

Looking back at traditional Computer Graphics, billboards have been a common staple for rendering complicated, semi-transparent features like leaves, grass, or particle effects in general.
We want to highlight here the similarities between 2DGS and billboards: in 2DGS, primitives are oriented 2D disks with a position, orientation and size in 3D, a constant color and opacity defined by a Gaussian distribution. Billboards are typically 2D squares with a position, orientation and size in 3D and a spatially-varying color and opacity using texture mapping.

To combine the best of both representations, we present \textit{Gaussian Billboards}, an extension to 2DGS in which the color of the splat is spatially modulated using a small per-primitive texture, see \cref{fig:teaser}a.
We show that replacing the per-primitive solid color with a per-primitive texture improves the reconstruction capability when a fixed number of primitives is employed (see \cref{fig:teaser}b), and also when the number of primitives is changed over the course of the optimization with densification and pruning strategies (see \cref{fig:teaser}c).
In short, our contributions are:
\begin{enumerate}
    \item An extension to 2DGS with a spatially-varying color defined by a per-primitive texture,
    \item A detailed ablation of the introduced hyperparameters,
    \item An evaluation of 2D and 3D scene representations showing the improved capabilities of Gaussian Billboards in representing texture details.
\end{enumerate}
It is worth mentioning that, while these contributions are novel, concurrent work by Rong~\etal~\cite{Rong2024-yq} named \textit{GStex} also explores similar ideas to enhance 2DGS with spatially varying textures.

\section{Related Work}
\label{sec:relwork}

We envision the proposed \textit{Gaussian Billboards} as a method that is orthogonal to other techniques for improving the rendering quality of 3DGS and 2DGS.

Yu~\etal~\cite{Yu2023-dq} and Yan~\etal~\cite{Yan2023-ul} show how to overcome aliasing errors during rasterization if Gaussian splats become very small from far-away cameras.
Improvements to the densification, splitting and training heuristics were presented by Zhang~\etal~\cite{Zhang2024-tt}, Ye~\etal~\cite{Ye2024-db} and Fang\&Wang~\cite{Fang2024-ug}. Especially in the context of 3DGS, the volumetric splats are approximated as flat disks and then sorted by the depth of the center position for alpha blending. This approximation can lead to popping artifacts due to inconsistent sorting between views. Radl~\cite{Radl2024-mn} show how to remedy these rendering errors using per-pixel sorting.
If the input views are of particular low resolution, Feng~\etal~\cite{Feng2024-eq} show how to utilize sub-pixel constraints and 2D super-resolution networks to recover sharp details. Similarly, Seiskari~\etal~\cite{Seiskari2024-ss} show how to explicitly compensate for motion blur and rolling shutters if the input images come from a video sequence. Alternatives to the usually used Adam optimizer have been explored by H\"ollein~\etal~\cite{Hollein2024-ca}.

More closely related to our approach is 3D Half-Gaussian Splatting by Li~\etal~\cite{Li2024-nv} where the 3D volumetric primitive is split into two halves and a separate color and opacity can be assigned to each half.
As one of the few works explicitly exploiting the $uv$-parametrization of 2DGS, Yu~\etal~\cite{Yu2024-gv} show how to modulate the Gaussian distribution of opacity with additional Hermite splines for sharper opacity borders.
Texture-GS by Xu~\etal~\cite{Xu2024-sx} operates on 3DGS and utilizes a small MLP to perform a global UV-unwrapping that maps the 3D splat position to a uv position of a global texture atlas. Due to the smooth mapping, this method facilitates simple texture edits in texture space, but due to its global nature it is limited to simple, isolated objects.

Concurrent work called GStex by Rong~\etal~\cite{Rong2024-yq} also explores per-primitive color textures. Differences between GStex and our proposed method include that we assign a small texture to each primitive right from the start that is optimized from the beginning, whereas GStex first trains with per-primitive solid colors and only adds textures in later epochs with potentially different resolutions per primitive. Furthermore, we conduct a detailed ablation of the spatial extent of the texture within the primitive and a different optimum compared to the value reported by GStex.



\section{Method}
\label{sec:method}
Before we present the proposed \textit{Gaussian Billboards} method in \cref{sec:method:billboards}, we first revisit the foundations of 3D and 2D Gaussian Splatting.

\subsection{Gaussian Splatting Fundamentals}

In the seminal work by Kerbel~\etal~\cite{Kerbl2023-cv}, 3D Gaussian Splatting (\textit{3DGS}) was introduced as a faster, higher quality 3D scene representation, compared to, \eg, Neural Radiance Fields (NeRFs)~\cite{Mildenhall2020-mp}.
Instead of representing the scene as a continuous, colored volume stored in an implicit neural network, 3DGS represents the scene as a collection of 3D Gaussian primitives that can be efficiently rasterized (\textit{splatted}) to a 2D screen~\cite{Zwicker2001-pc,Zwicker2002-xh}.
In 3DGS, a single primitive $k$ is defined by its location in 3D $\mathbf{p}_k$, its covariance matrix $\mathbf{\Sigma}_k$ parameterized via a scale $\mathbf{s}_k$ and rotation quaternion $\mathbf{q}_k$, an RGB color $\mathbf{c}_k$, and an opacity $\alpha_k$.
This representation can be viewed as an oriented and colored ellipsoid. To make this representation differentiable and optimizable, 3DGS turns these ellipsoids into 3D Gaussian primitives by multiplying the opacity with a 3D Gaussian distribution~\cite{Kerbl2023-cv}:
\begin{equation}
    \mathcal{G}(\mathbf{p}) = \text{exp}\left(-\frac{1}{2}(\mathbf{p}-\mathbf{p}_k)^T \mathbf{\Sigma}^{-1}(\mathbf{p}-\mathbf{p}_k)\right) .
\end{equation}
Using the techniques by Zwicker~\etal~\cite{Zwicker2001-pc}, the perspective projection of a 3D Gaussian distribution can be analytically approximated giving rise to a 2D Gaussian distribution $\mathcal{G}^{2D}$ in screen space.
Then, all splats intersecting the camera ray of a given pixel $\mathbf{x}$ are blended together using front-to-back alpha compositing.
\begin{equation}
    \mathbf{c}(\mathbf{x}) = \sum_{k=1}^K \mathbf{c}_k \alpha_k \mathcal{G}_k^{2D}(\mathbf{x}) \prod_{j=1}^{k-1}(1 - \alpha_j \mathcal{G}_j^{2D}(\mathbf{x})) .
\end{equation}

2D Gaussian Splatting~\cite{Huang2024-2dgs} simplifies the above method by representing the scene as a collection of oriented, flat 2D primitives for better view consistency and improved surface reconstruction.
Here, the geometry of a 2D Gaussian primitive is defined by a center position $\mathbf{p}_k \in \mathbb{R}^3$, two orthogonal tangent vectors $\mathbf{t}_{u,k}, \mathbf{t}_{v,k} \in \mathbb{R}^3$ (derived from a rotation quaternion) and a 2D scaling vector $S = (s_u, s_v) \in \mathbb{R}^2$.
This defines a local tangent plane parameterized by $\mathbf{u}=(u,v)$:
\begin{equation}
    P(\mathbf{u}) = \mathbf{p}_k + s_u \mathbf{t}_{u,k} u + s_v \mathbf{t}_{v,k} v .
    \label{eq:2dgs-plane}
\end{equation}
The 2D Gaussian distribution for modulating the opacity is now equally defined in this $uv$-space using
\begin{equation}
    \mathcal{G}(\mathbf{u}) = \text{exp}\left(-\frac{1}{2}(u^2+v^2)\right) .
\end{equation}

\subsection{Gaussian Billboards}
\label{sec:method:billboards}

In the original 2DGS implementation~\cite{Huang2024-2dgs}, each 2D primitive is rendered with a constant color $\mathbf{c}_k$ across the entire 2D plane it spans. This ties both geometry and textural information together in the same representation, limited by the same quantity, \ie the number of splats.

We observe, however, that in most scenes, the texture information is higher than geometry information.
Therefore, we take inspiration from textured billboards of traditional Computer Graphics and introduce a spatially-varying color representation by bi-linearly interpolating a color grid based on the $uv$-parametrization of the 2D primitive, see \cref{eq:2dgs-plane}.
This enables modelling textures better with the same or fewer number primitives.
Let $N\in\mathbb{N}$ the resolution of the per-primitive color grid. Since the Gaussian primitive has infinite support, we have to define a region in the $uv$-plane in which the texture is defined. Let $\sigma\in\mathbb{R}^+$ denote said spatial extent, then the texture grid is contained within $[-\sigma,+\sigma]^2$ in the $uv$-plane.
The pixel color of the current primitive is then given by
\begin{equation}
    \mathbf{c}(u,v)_k = \text{bilinear}(C_k, \frac{u}{\sigma}, \frac{v}{\sigma}) ,
\end{equation}
where $\text{bilinear}(C, x, y)$ represents bilinear interpolation into a grid $C$ of colors with shape $N \times N$, evaluated at normalized coordinates $x,y \in [-1,+1]$.
Since the color grid is defined in the $uv$-parametrization, it naturally scales with $S$ and rotates with $\mathbf{t}_{u,k}, \mathbf{t}_{v,k}$, see \cref{eq:2dgs-plane}.

\begin{figure}
    \centering
    \includegraphics[width=0.60\linewidth]{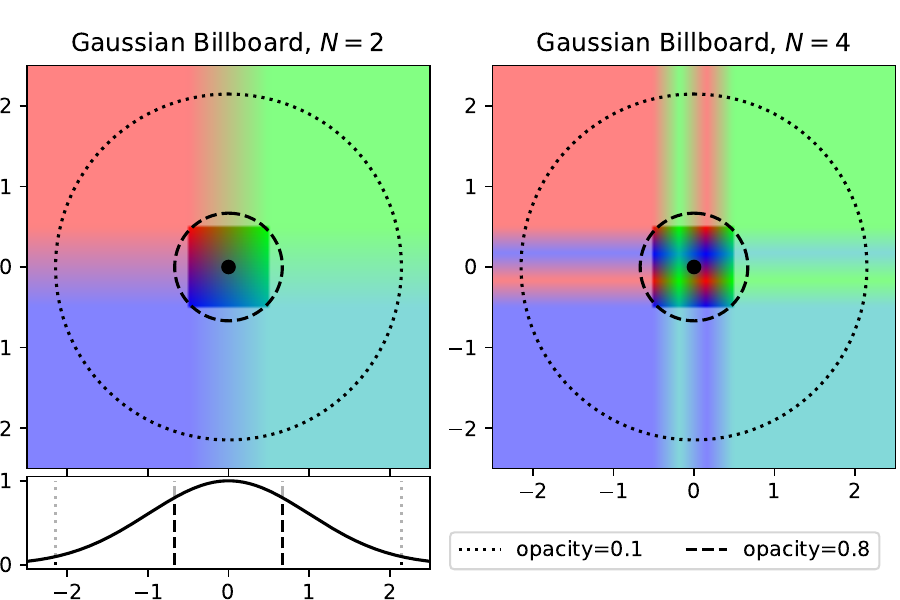}%
    \includegraphics[width=0.38\linewidth]{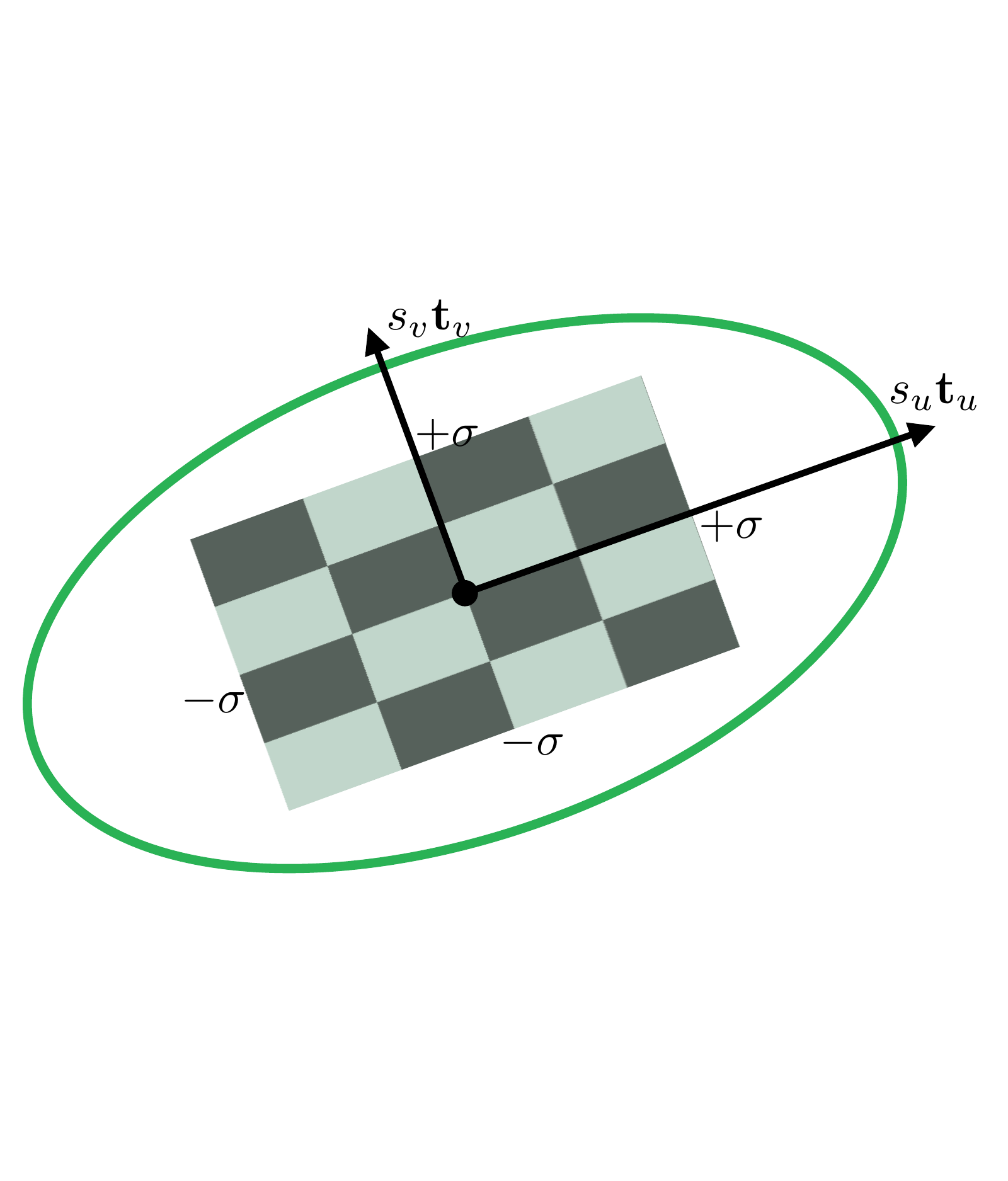}
    \caption{
        The color grid of Gaussian Billboards spans the $uv$-parametrization of the 2D primitive in the range of $[-\sigma,+\sigma]$ with $\sigma=0.5$. This way, the border of the grid where the interpolation starts being clamped roughly corresponds to the isoline of $0.8$ opacity. On the right you can see how the texture within a primitive is rotated and scaled with $s_u,s_v,\mathbf{t}_u,\mathbf{t}_v$.
    }
    \label{fig:visgrid}
\end{figure}

A visualization of the proposed spatial color can be seen in \cref{fig:visgrid} for $N=2,4,8$ with a red-green-blue checkerboard texture.
The darker area indicates the area of $[-\sigma,+\sigma]^2$. However, since the Gaussian primitive has infinite support, we border-clamp the $uv$-values in the bilinear interpolation function. The result of this clamping is visualized in the brighter shaded region.
An intuitive choice for $\sigma$ is to pick it such that the texture grid covers the majority of the visible are of a primitive to minimize the area where this clamping occurs, \eg, at the isoline of $0.1$ opacity (dotted line) with $\sigma=2.0$.
We found, however, that the value for $\sigma$ has a significant impact on how much the proposed Gaussian Billboards can improve over traditional 2DGS. An ablation of this value can be found in \cref{sec:results:image}, with the optimal value being $\sigma=0.5$, or the full color grid residing within the isoline of $0.8$ opacity (dotted line), shown in \cref{fig:visgrid}.
For all of our results, if not otherwise mentioned, we use a value of $\sigma=0.5$.

\subsection{Training}
\label{sec:method:training}
For training the scene representation, we follow the standard practices of 3DGS and 2DGS. We initialize the primitives using structure-from-motion if available and add slight color variations to the color grid. We use Adam~\cite{Kingma2014-fk} as the optimizer and train only with a photometric loss.
Since we envision \textit{Gaussian Billboards} as an orthogonal method to other improvement techniques, we opted to keep all other training hyperparameters at the default values used in traditional 2DGS to isolate the effects of introducing spatially-varying colors.

For optimal performance, we integrated Gaussian Billboards directly into the fused CUDA kernels of \textit{gsplat}~\cite{ye2024gsplatopensourcelibrarygaussian} with gradients being propagated to all parameters, including the color grid and the $uv$ coordinate, thus also to the position, scale and orientation of the primitive.

To reduce global memory bandwidth in the forward and backward pass of the rasterization kernels, \textit{gsplat} first loads and accumulates the per-primitive parameters and their gradients in shared memory for a batch of pixels and primitives called a tile, before writing them out to global memory.
Shared memory is bound by hardware to 32kB to 100kB depending on the GPU architecture
and sets an upper bound of the per-primitive texture resolution.
For $N=2$ and $N=4$, the additional parameters fit into shared memory without changes. For $N=8$, however, the required shared memory exceeds the available memory in hardware (tested on an A6000 and RTX3090), and the tile size must be reduced from $16 \times 16$ to $8 \times 8$, drastically increasing the rendering and training times (see \cref{sec:results:image} and \cref{fig:image-texture-resolution}).
This also makes larger resolutions of, \eg, $N=16$ infeasible.


\section{Results}
\label{sec:results}

We first evaluate the proposed method and design decisions on the task of fitting a 2D image, see \cref{sec:results:image}.
Then we qualitatively and quantitatively evaluate Gaussian Billboards on a range of 3D scene reconstruction tasks using the NeRF360 dataset~\cite{Barron2021-lg} and human faces, see \cref{sec:results:3d}.

\subsection{Ablations on Image Overfitting}
\label{sec:results:image}

\begin{figure}[tbp]
    \centering
    \includegraphics[height=3cm]{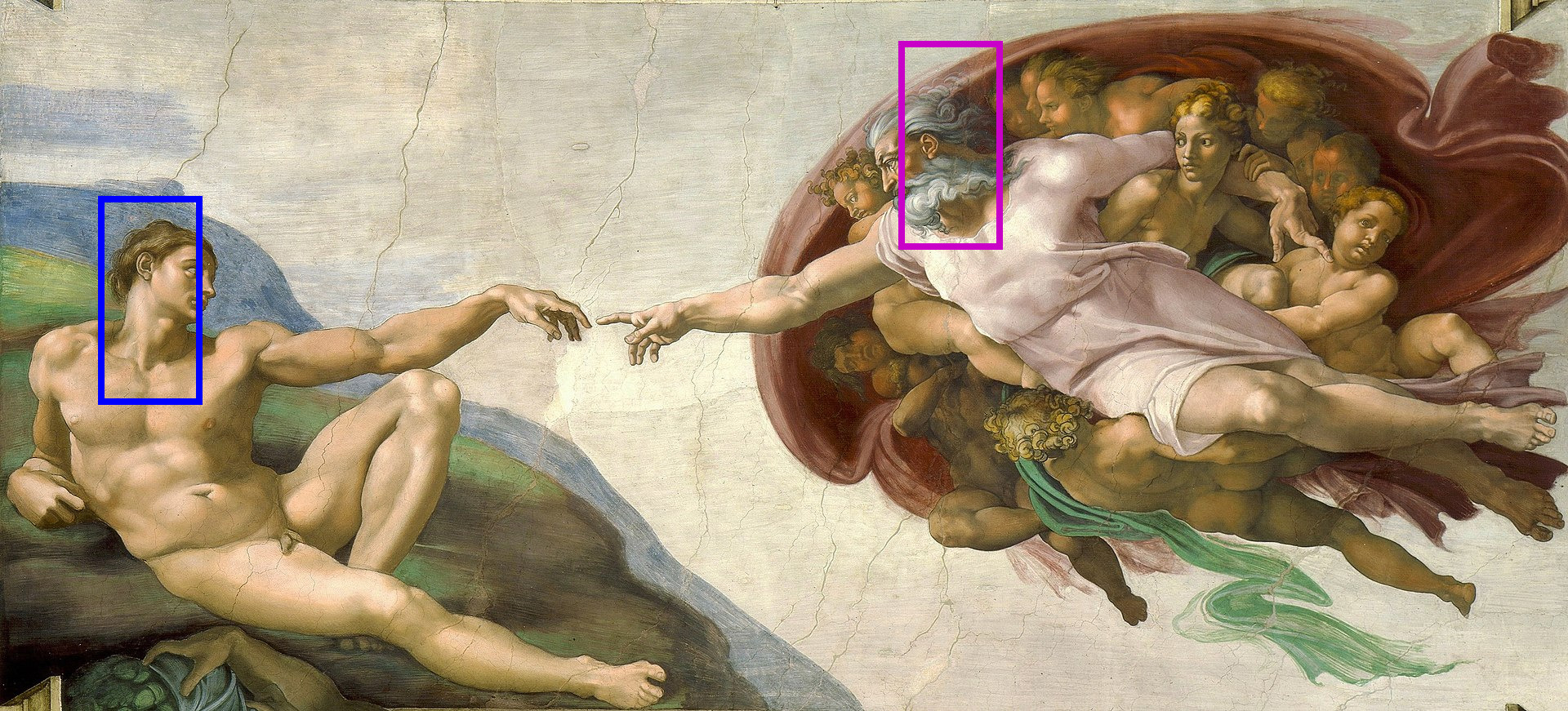}\\%
    \includegraphics[height=3cm]{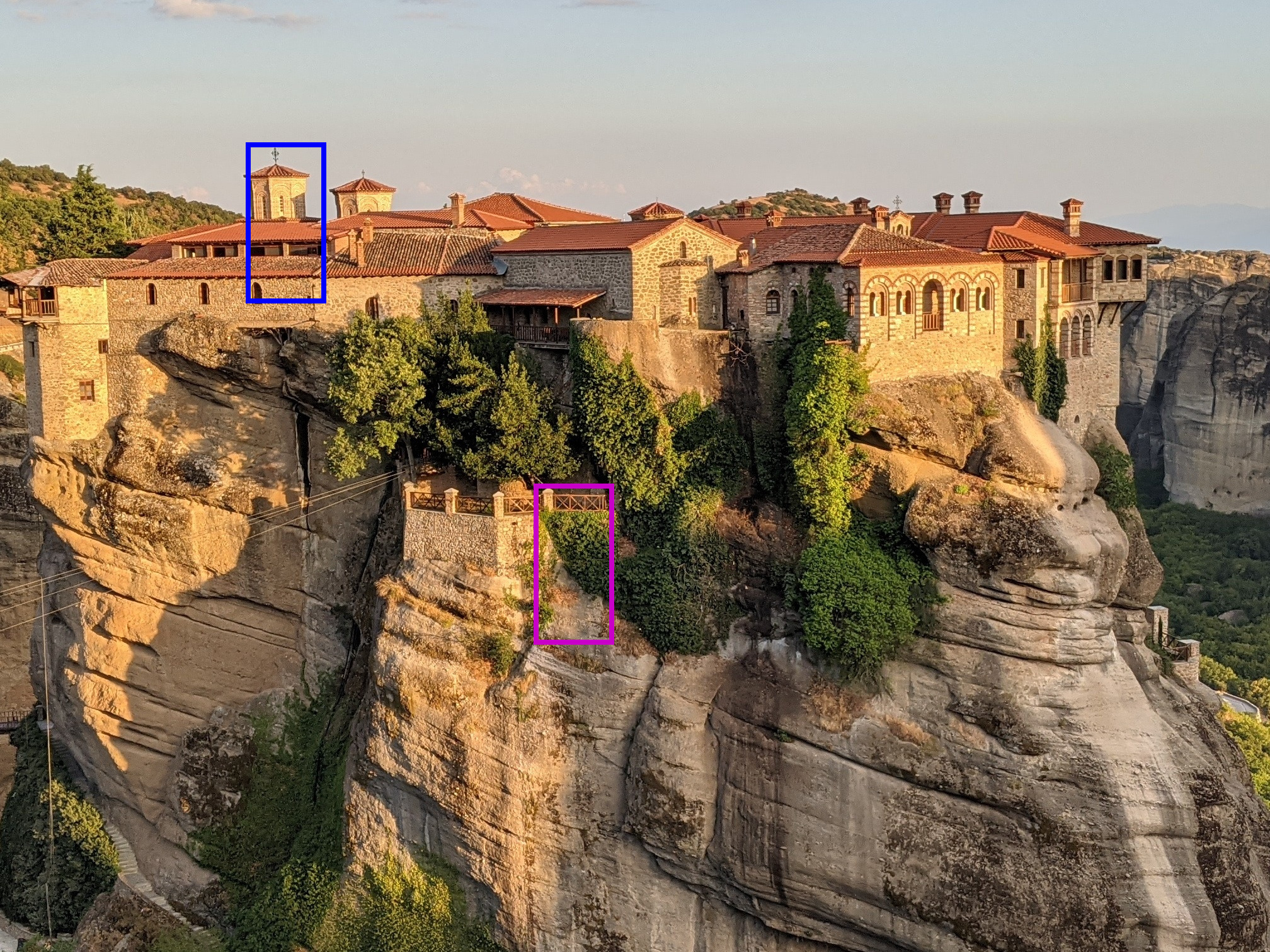}~%
    \includegraphics[height=3cm]{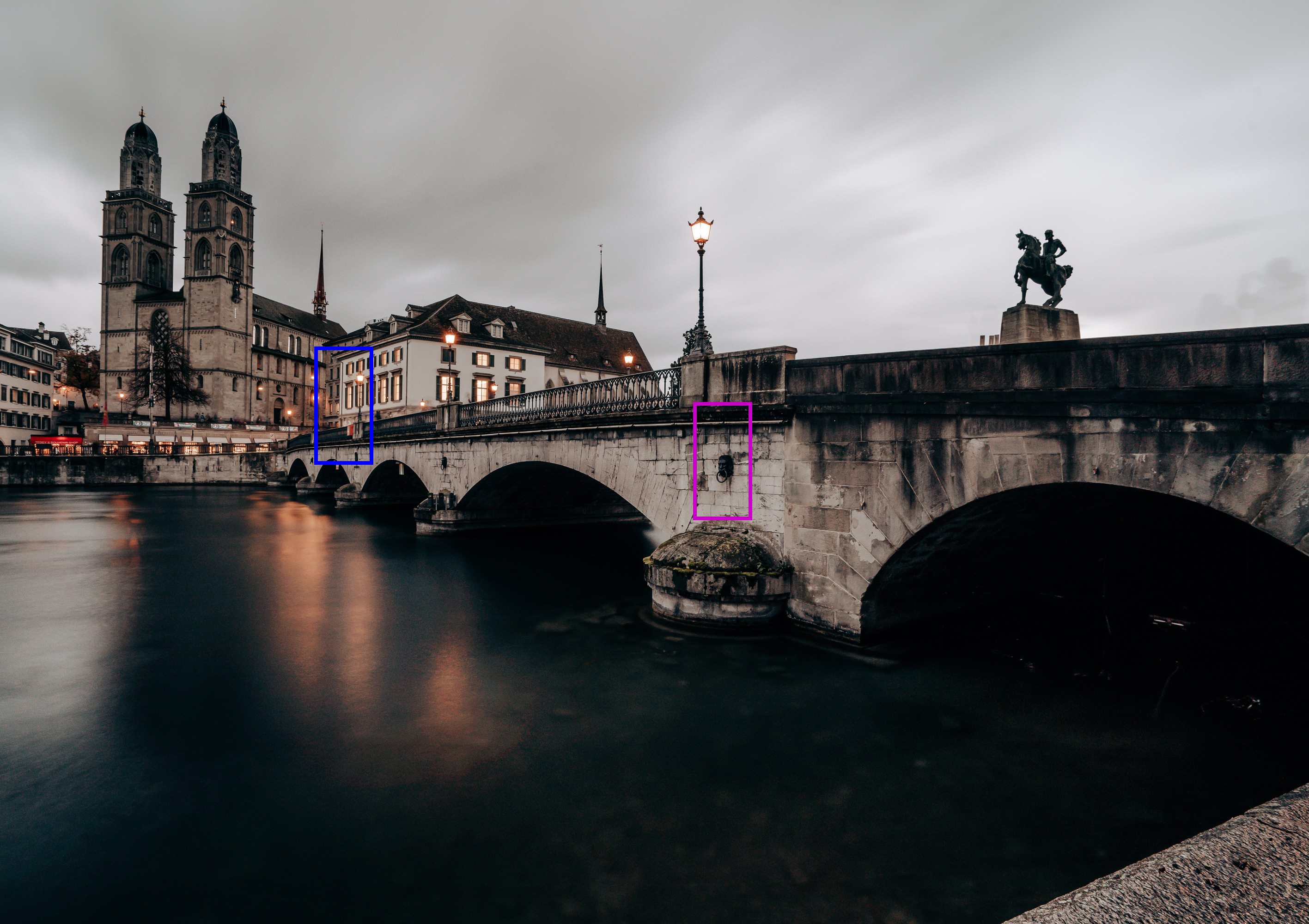}%
    \caption{The three images used for image fitting tests and hyperparameter ablations: \textit{Creation of Adam}, \textit{Meteora}, and \textit{Zurich}. The marked areas indicate the crops visualized in \cref{fig:image-texture-resolution}.}
    \label{fig:results:input-images}
\end{figure}

For overfitting a single image, we define one camera focussed on the image in +Z direction and restrict the rotation of the primitives to only rotate around the Z-axis. This way, the plane of the 2D primitives aligns with the image plane. The primitives are initialized with random positions in the image plane, random orientation, size, and color.
We perform these tests on three images (see \cref{fig:results:input-images}), \textit{Creation of Adam} by Michelangelo of size $1920 \times 871$\footnote{\textit{Creation of Adam} is taken from Wikicommons, public domain.},
a photograph of the \textit{Meteora} monastery in Greece with resolution $2016 \times 1512$, and a photograph of the M\"unsterbr\"ucke in \textit{Zurich}, Switzerland, with resolution $1833 \times 2000$. All results shown are trained for 20000 epochs with an MSE loss.

\begin{figure}[tbp]
    \centering
    \begin{overpic}[width=0.47\linewidth]{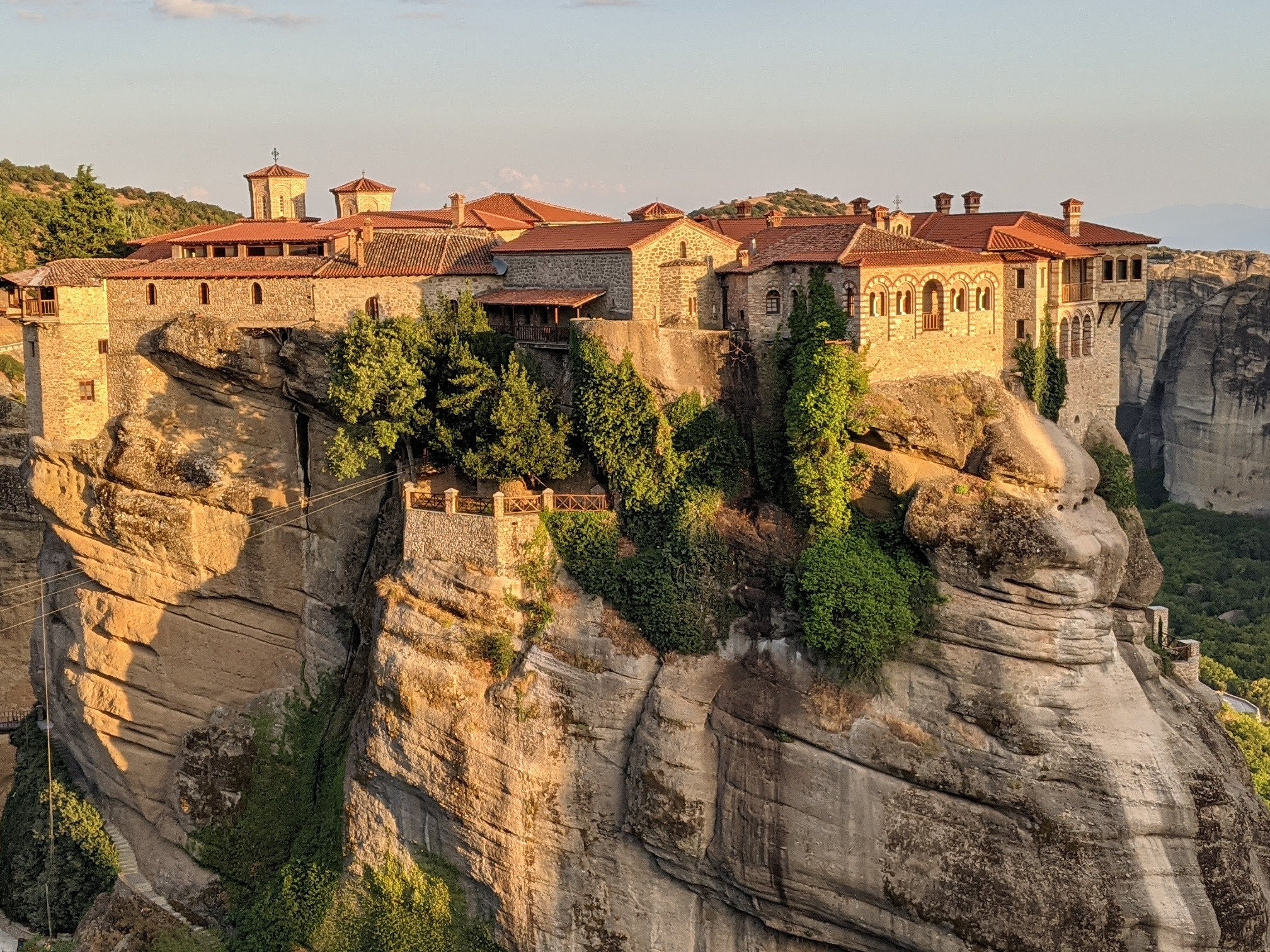}
        \put(0,3){\footnotesize\colorbox{white}{a) Target}}
    \end{overpic}\\[2pt]%
    \begin{overpic}[width=0.47\linewidth]{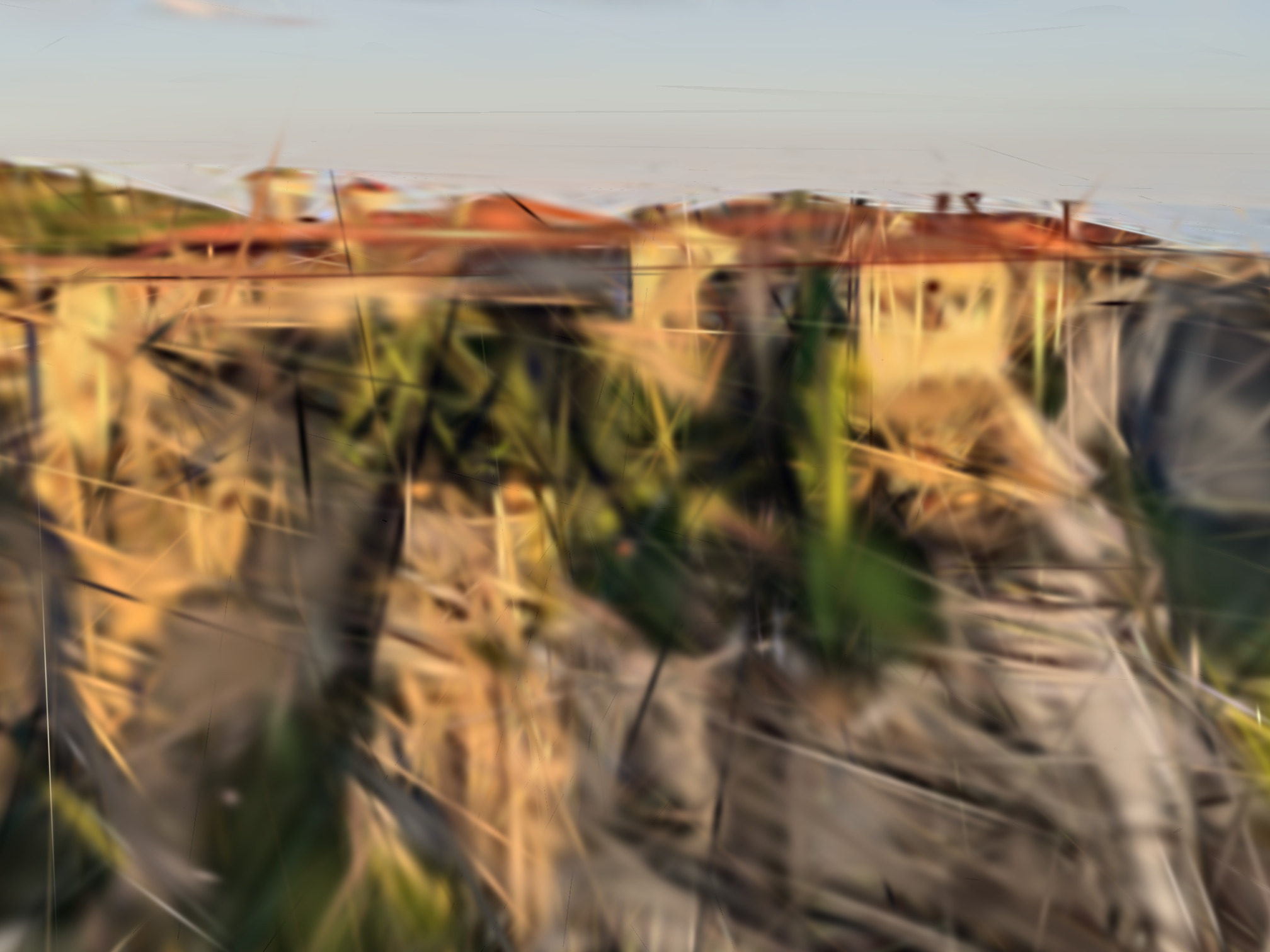}
        \put(0,3){\footnotesize\colorbox{white}{b) 2DGS}}
    \end{overpic}~~%
    \begin{overpic}[width=0.47\linewidth]{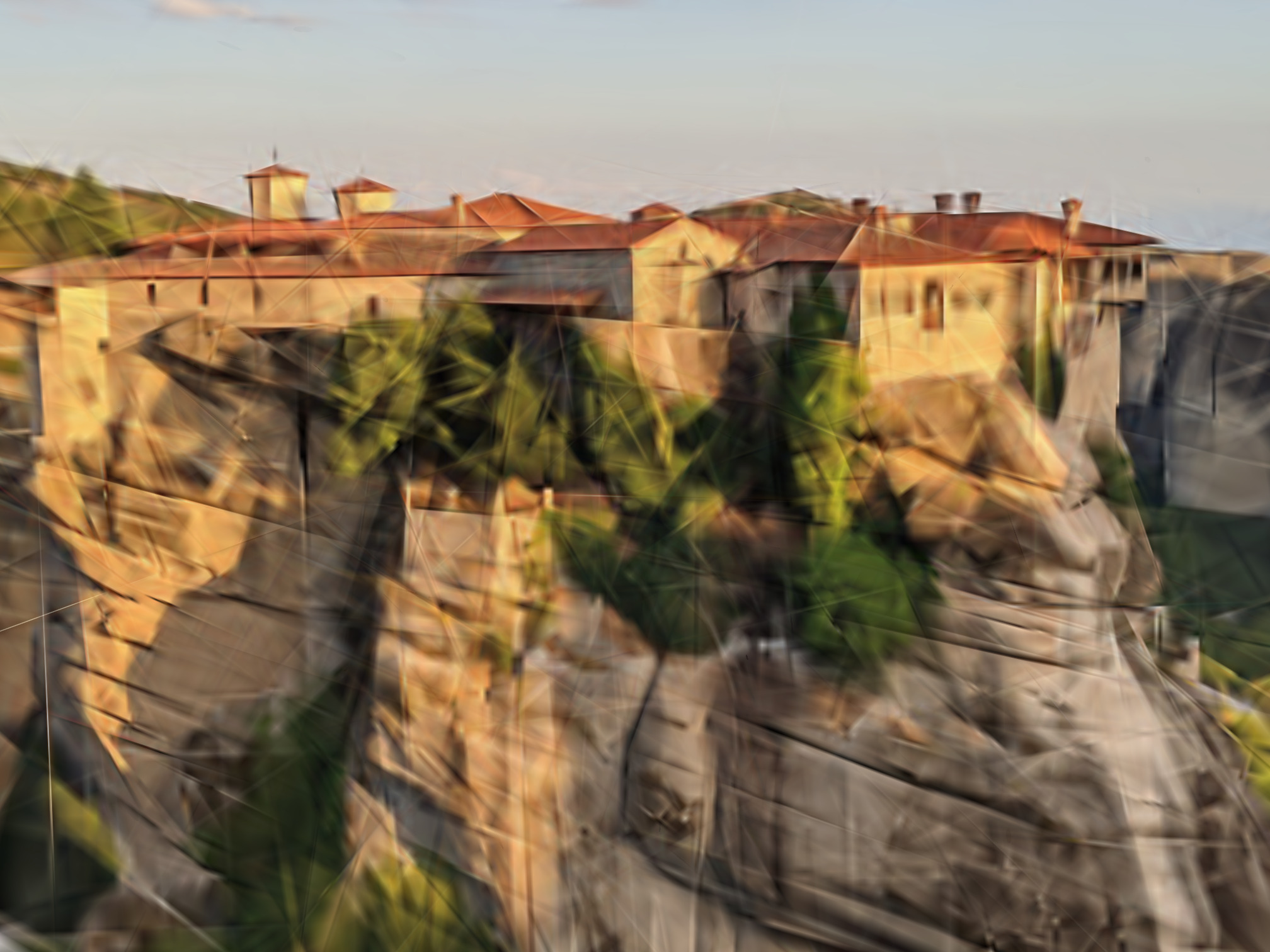}
        \put(0,3){\footnotesize\colorbox{white}{c) Ours $N=4$}}
    \end{overpic}\\[-5pt]%
    ~\hfill {\tiny PSNR: 19.8, SSIM: 0.380, LPIPS: 0.796} \hfill \hfill {\tiny PSNR: \textbf{20.7}, SSIM: \textbf{0.414}, LPIPS: \textbf{0.691}} \hfill ~\\[4pt]%
    \begin{overpic}[width=0.47\linewidth]{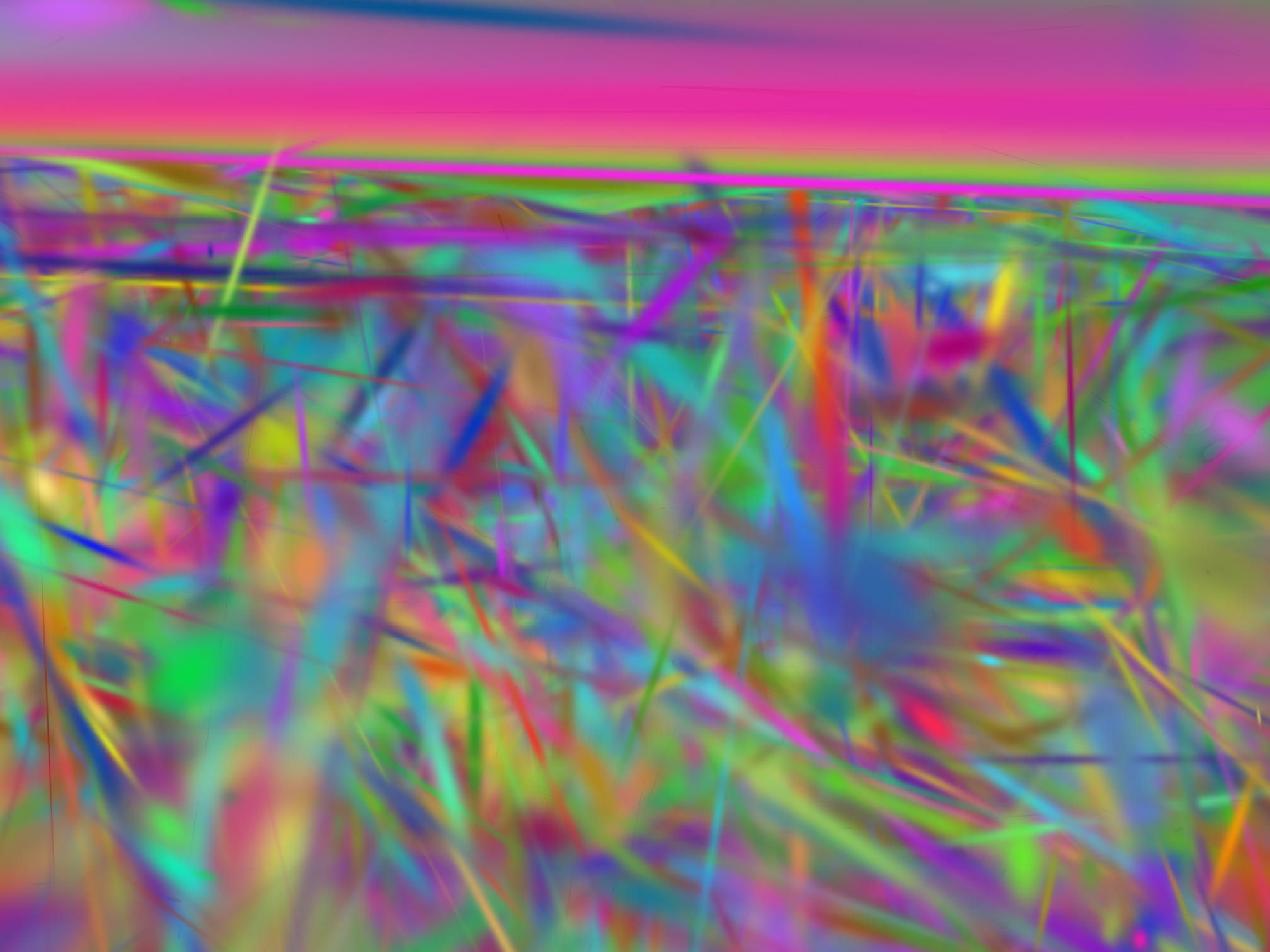}
        \put(0,3){\footnotesize\colorbox{white}{d) 2DGS, random color}}
    \end{overpic}~~%
    \begin{overpic}[width=0.47\linewidth]{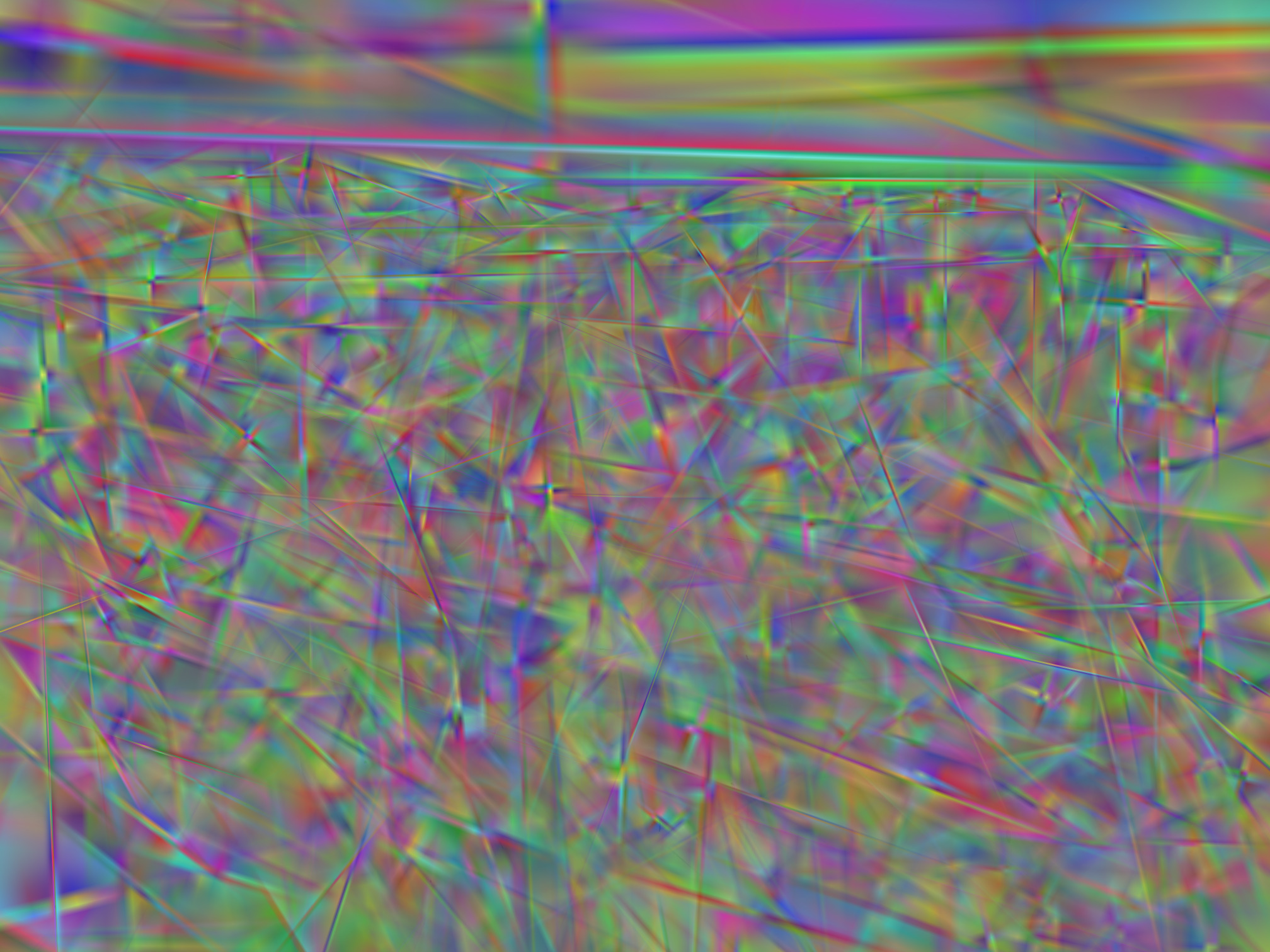}
        \put(0,3){\footnotesize\colorbox{white}{e) Ours $N=4$, random colors}}
    \end{overpic}\\[2pt]%
    \begin{overpic}[width=0.47\linewidth]{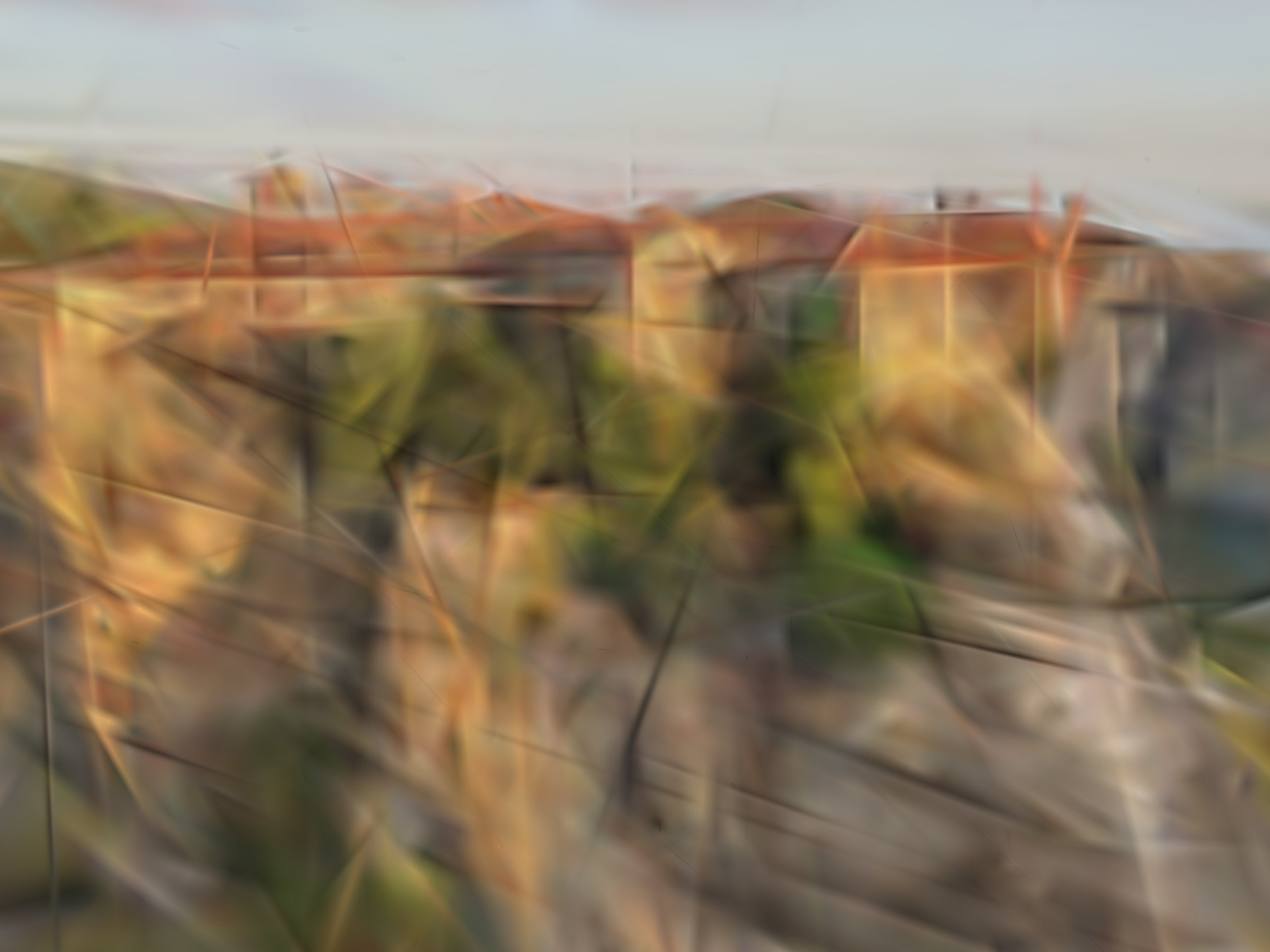}
        \put(0,3){\footnotesize\colorbox{white}{f) Ours $N=4$, mip level 1}}
    \end{overpic}~~%
    \begin{overpic}[width=0.47\linewidth]{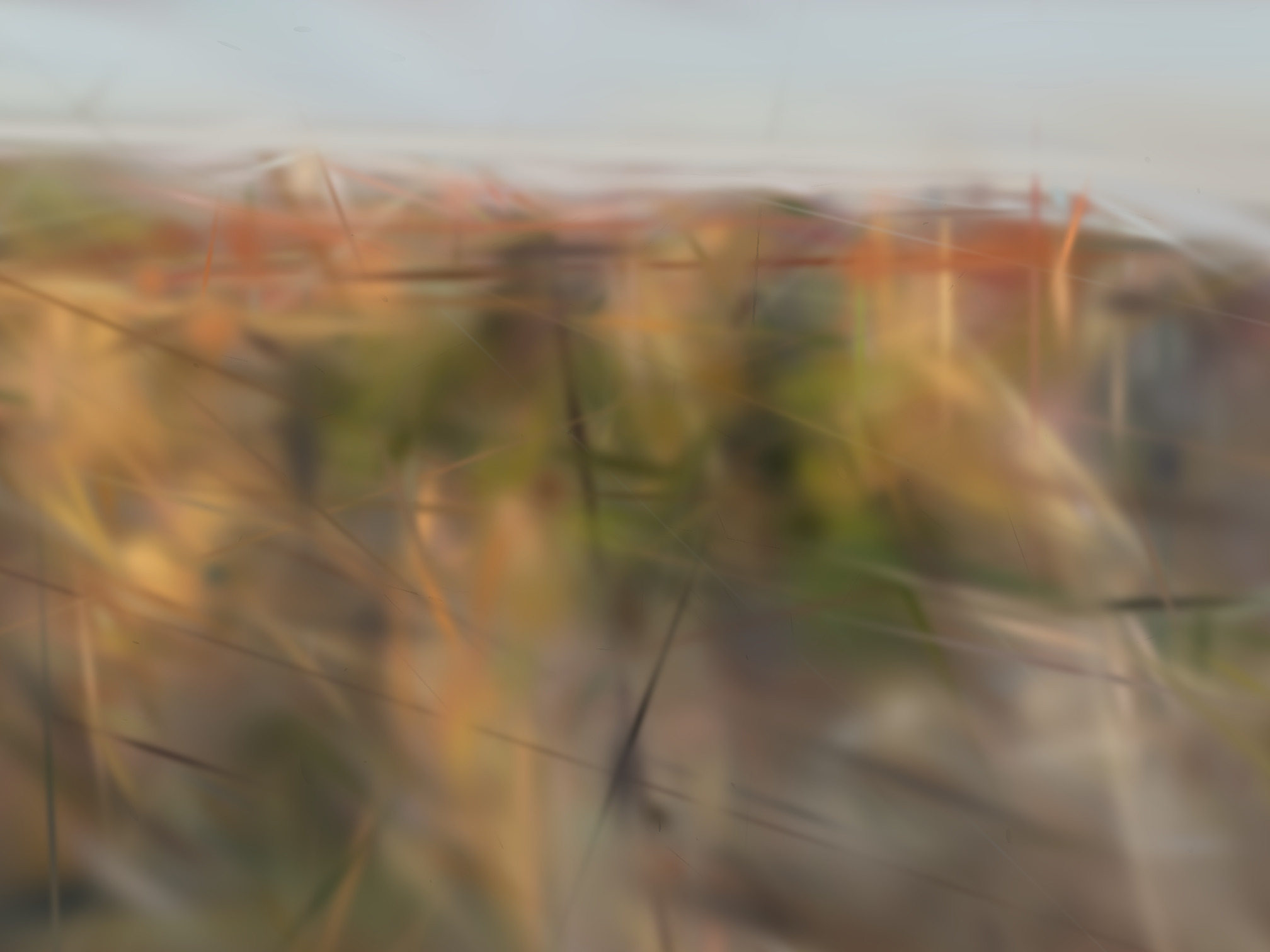}
        \put(0,3){\footnotesize\colorbox{white}{g) Ours $N=4$, mip level 2}}
    \end{overpic}%
    \caption{
        Overfitting a single image (a) with a fixed number of 1000 splats.
        For that limited number of splats, Ours with $N=4$ (c) achieves better quality than Traditional 2DGS (b) in terms of quantitative statistics and image sharpness. If we assign random colors, the individual splats and the bilinear interpolation of the color grid become visible (d,e). To verify that the optimization utilizes the additional color information, we downsample the color grid to a resolution of $2 \times 2$ (f) and a single color (g).
    }
    \label{fig:result:adam1000}
\end{figure}

In a first test, we fix the number of splats to 1000 and compare traditional 2DGS with our proposed method, see \cref{fig:result:adam1000}, and $N=4$. For this limited number of splats, our method achieves visually a sharper and more texture-rich image (c vs b) and improved quantitative statistics.
To visualize the individual splats, we assign random colors to each splat (d and e).
Finally, to verify that the optimization utilizes the additional color parameters, we render the image with the color grid downsampled to a size of $2 \times 2$ (f) or a single color (g) using a box filter. As one can see, the image quality drastically decreases, highlighting that the color grid is now crucial for the image quality.



\paragraph{Ablation of the Grid Extent $\sigma$.}

\begin{figure}[tbp]
    \centering
    \includegraphics[width=\linewidth]{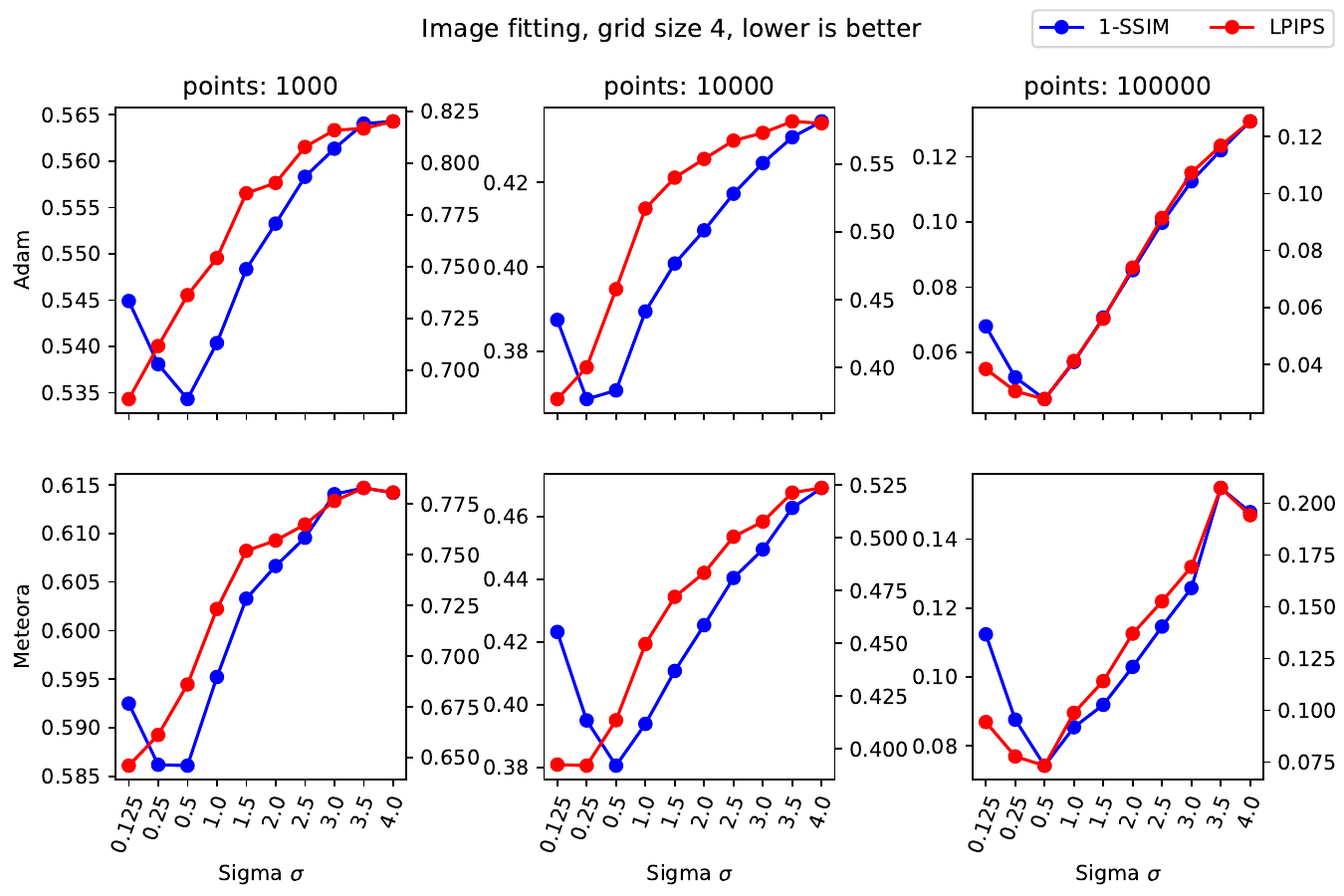}\\\vspace{-1.5em}%
    \begin{overpic}[width=0.24\linewidth,trim={40 0 80 0},clip]{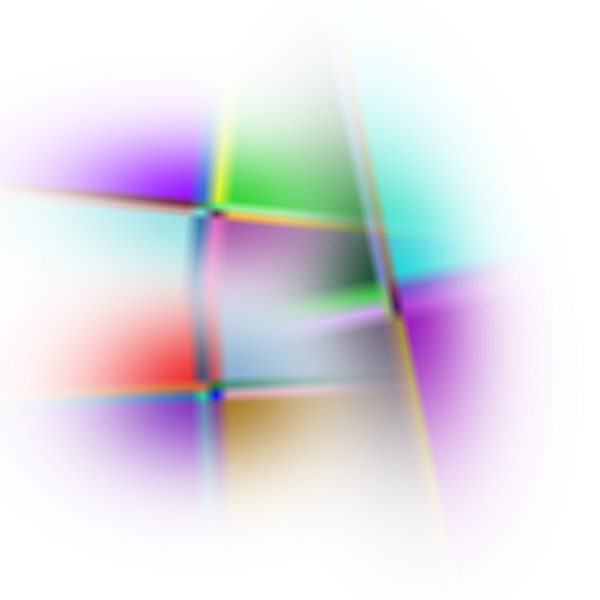}
        \put(2,5) {$\sigma=0.25$}
    \end{overpic}
    \begin{overpic}[width=0.24\linewidth,trim={40 0 80 0},clip]{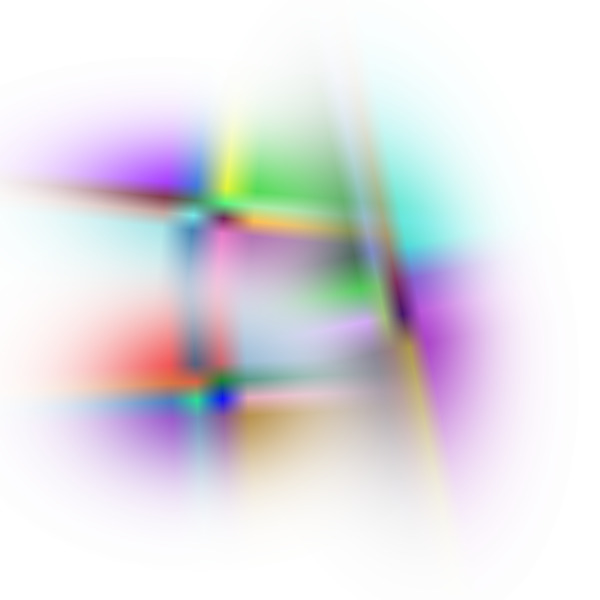}
        \put(2,5) {$\sigma=0.5$}
    \end{overpic}
    \begin{overpic}[width=0.24\linewidth,trim={40 0 80 0},clip]{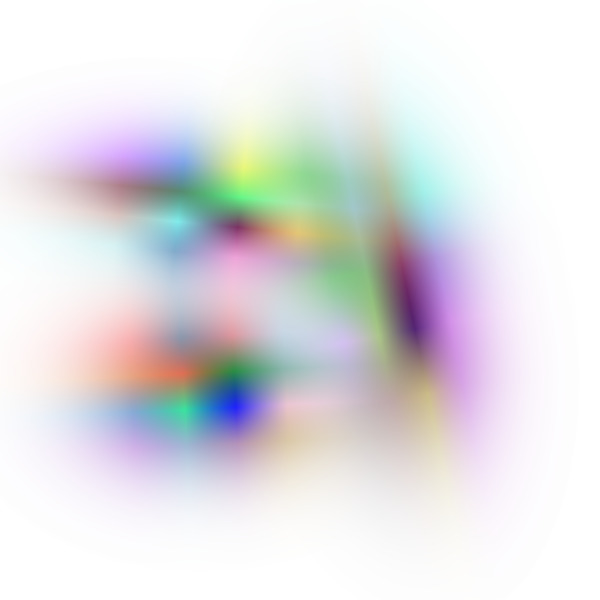}
        \put(2,5) {$\sigma=1.0$}
    \end{overpic}
    \begin{overpic}[width=0.24\linewidth,trim={40 0 80 0},clip]{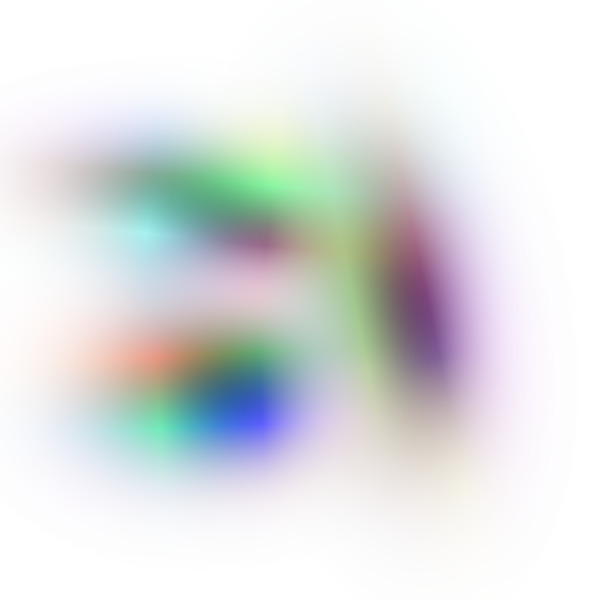}
        \put(2,5) {$\sigma=2.0$}
    \end{overpic}
    \caption{Ablation of the spatial extent of the grid in the $uv$-plane, $\sigma$, for overfitting of \textit{Creation of Adam} and \textit{Meteora} (see \cref{fig:results:input-images}) with $N=4$. With a lower value for $\sigma$, the texture information is concentrated in the area of the primitive with the highest opacity. As can be seen in the plot, the optimal value for $\sigma$ is achieved at around $0.5$. For a visualization of what different values for $\sigma$ mean, the bottom row shows three Gaussian primitives with a random texture rendered with four values for $\sigma$.}
    \label{fig:ablation:sigma}
\end{figure}
Next, we evaluate the influence of the hyperparameter $\sigma$, the spatial extent of the grid in the $uv$-parametrization, on an image fitting task with a fixed number of primitives (see \cref{fig:ablation:sigma}).
A lower value for $\sigma$ indicates that the texture details are concentrated in the center area of the Gaussian primitive where the opacity is highest. As can be seen in the statistics of \cref{fig:ablation:sigma}, across two different test images and different number of primitives, the reconstruction quality measured in SSIM~\cite{Wang2004-bn} and LPIPS~\cite{Zhang2018-pz} scores, improves with lower values for $\sigma$.
This reaches an optimal point of around $\sigma=0.5$.
We argue that for even smaller values, the texture essentially collapses to a binary quadrant image and cannot make use of the full resolution of a $4 \times 4$ grid in this case anymore.

\paragraph{Ablation of the Texture Resolution $N$.}

\begin{figure*}
    \centering
    \includegraphics[width=0.9\textwidth]{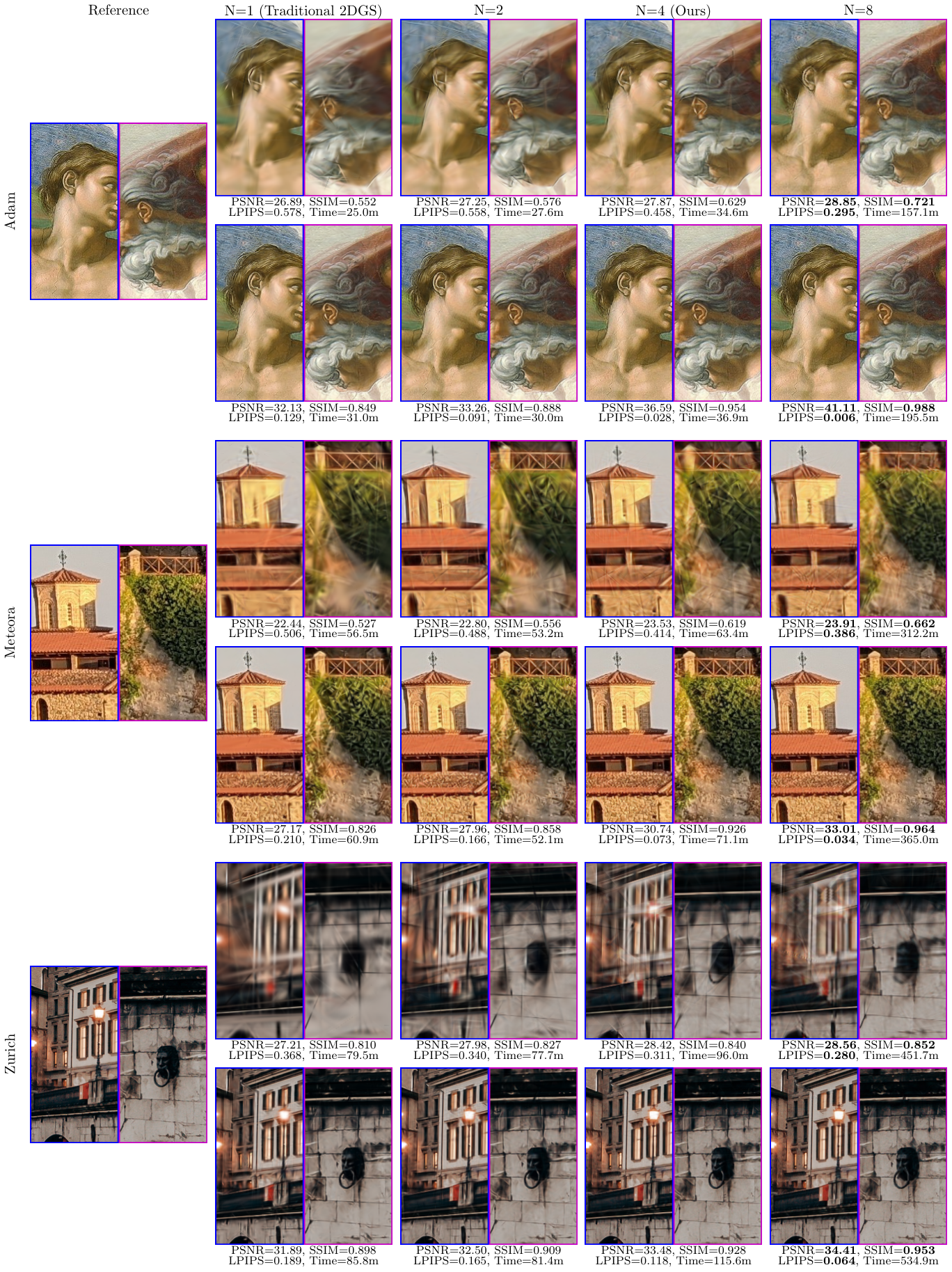}
    \caption{Evaluation of the texture resolution $N$ for image fitting with a fixed number of primitives, per scene with 10,000 primitives (top row) and 100,000 primitives (bottom row). With an increase in texture resolution, the Gaussian Billboards consistently lead to better fittings.}
    \label{fig:image-texture-resolution}
\end{figure*}
With a fixed value of $\sigma=0.5$ we now turn to analyzing the texture resolution $N$ and its impact on the reconstruction quality and training speed compared to traditional 2DGS without spatially varying colors.
We perform this experiment on all three test images with a fixed number of primitives of 10,000 and 100,000.  See \cref{fig:image-texture-resolution} for the results.

Qualitatively, the difference can be seen best for 10,000 primitives (upper row of \cref{fig:image-texture-resolution} per image).
With $N=1$ (traditional 2DGS), many details are either blurred out in \textit{Creation of Adam} and \textit{Zurich} or approximated with elongated splats in \textit{Meteora}.
From $N=2$ to $N=4$, the reconstructed image significantly improves. Texture detail of the paint becomes visible in \textit{Creation of Adam} and details in the plants and bricks become visible in \textit{Meteora} and \textit{Zurich}.
Continuing on to $N=8$ still shows improvements, but less noticeable.

Quantitatively, increasing the texture resolution strictly improves the PSNR, SSIM, and LPIPS scores, evaluated over the entire image.
For \textit{Creation of Adam} and 100,000 primitives, the SSIM score increases, for example, from $0.849$ for $N=1$ to $0.988$ for $N=8$, matching the input image almost pixel by pixel.

Enhancing the Gaussian primitives with spatially-varying textures, however, does come with some overhead in computation, but especially with an overhead of additional memory transfer and reduced tile size (see \cref{sec:method:training}) due to the additional parameters that must be considered in the splatting kernel.
We, therefore, look at the timings shown in \cref{fig:image-texture-resolution} now, all tests trained for 20000 epochs on an A6000 GPU.
For $N=2$ and $N=4$, the training time usually increases between 10\% and 30\% compared to traditional 2DGS. However, for $N=8$, the tile size has to be reduced and the training time increases dramatically for all test cases, \eg, by a factor of almost $4 \times$ for \textit{Meteora}.
Despite the improved quality with $N=8$, we found the increase in computation time to not be worth it. We, therefore, recommend $N=4$ and use that value for all following tests.

\subsection{Evaluation on 3D Scene Reconstruction}
\label{sec:results:3d}

In a 3D scene reconstruction task, the surface color can change depending on the view direction due to specular effects. We can model this as Spherical Harmonics as traditionally done in 3DGS and 2DGS~\cite{Kerbl2023-cv,Huang2024-2dgs} and directly extend this to a spatially varying color grid by storing $M$ spherical harmonics coefficients for each texel instead of a single RGB color.  This, however, quickly becomes expensive in the number of parameters for high spherical harmonics degrees.
Alternatively, we can treat the color grid as the albedo of the surface and then model the view-dependent appearance with explicit BSDF models~\cite{Jiang2023-ax}.
In this work, however, we opt to perform a diffuse-specular split. We use the proposed color grid as the view-independent Lambertian part and model the residual specular components as an additive term using a single set of spherical harmonics coefficients -- not spatially varying -- per primitive.

\begin{figure*}
    \centering
    \includegraphics[width=0.75\linewidth]{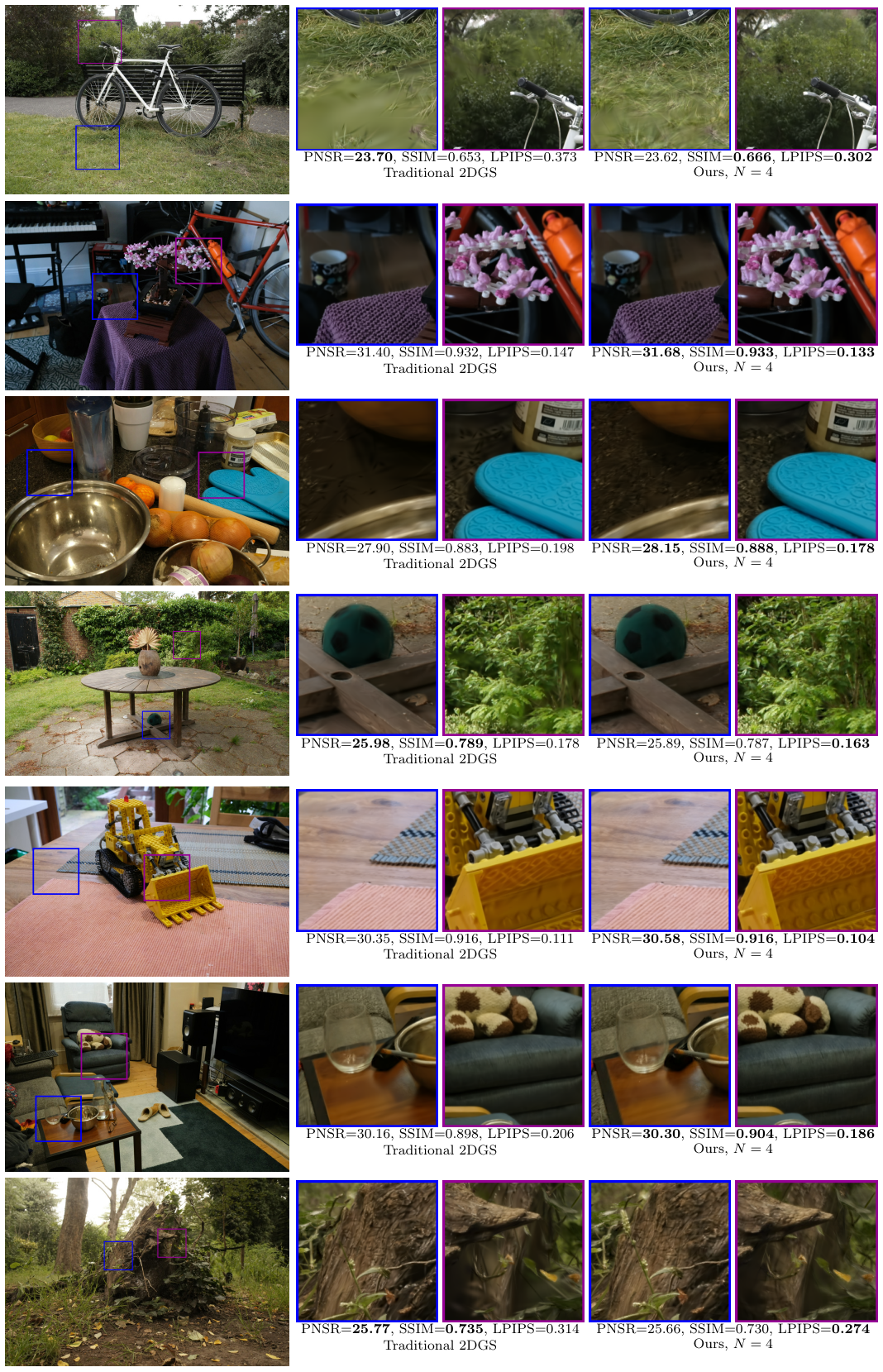}%
    \caption{Evaluation of traditional 2DGS compared to Gaussian Billboards (\textit{Ours}) on the NeRF360 validation images~\cite{Barron2021-lg}.  With the proposed method, areas that are seen from few images and areas that show texture variation with little 3D structure are reconstructed better, see \eg the grass and leaf areas in row 1 or the inside of the shovel of the toy bulldozer in row 5. The reported PSNR, SSIM and LPIPS statistics are averaged over all validation images. The left column shows the reference image. Images best viewed zoomed-in.}
    \label{fig:nerf360}
\end{figure*}

Results of the 3D scene reconstruction task on the NeRF360 dataset~\cite{Barron2021-lg}, downsampled by a factor of 2 to a resolution of approximately $2500 \times 1650$, can be found in \cref{fig:nerf360}. We train for 30000 epochs with default densification and pruning enabled, initialized with SfM points, and a loss of $M\!S\!E+0.2*S\!S\!I\!M$.
Quantitatively, our method ($N=4, \sigma=0.5$) consistently outperforms traditional 2DGS in terms of LPIPS statistics on validation images. For SSIM, \textit{Ours} outperforms \textit{2DGS} in 5 out of 7 cases with the two outliers being very close between both methods.

Qualitatively, we observe that \textit{Gaussian Billboards} improves the visual quality especially in areas where the geometry is fairly flat but shows a lot of texture variation. One case can be seen in the counter scene (third row) where the marble texture of the counter is represented better in \textit{Ours} (blue inset). Another example of the same behavior can be seen in the shovel of the toy bulldozer.
Even for areas of high spatial variation, \eg, the vegetation in the bicycle scene of row one, \textit{Ours} visually reconstructs more detail.

\paragraph*{Face Reconstruction.}

For a last test we evaluate \textit{Gaussian Billboards} on 3D face reconstruction from a multi-view camera setup with 14 training cameras arranged in a hemisphere around the front of the subject and one central test camera not used for training.
We use camera calibration with fiducial markers~\cite{Garrido-Jurado2014-pj} and a 3D face scan~\cite{Beeler2010-jo} to obtain a point cloud for initialization.  The background of the camera images with resolution $2047 \times 1535$ was removed using Lin \& Ryabtsev~\etal~\cite{Lin2021-rn}.
All training parameters remain the same as described above.

\begin{figure}[t]
    \centering
    \includegraphics[width=\linewidth]{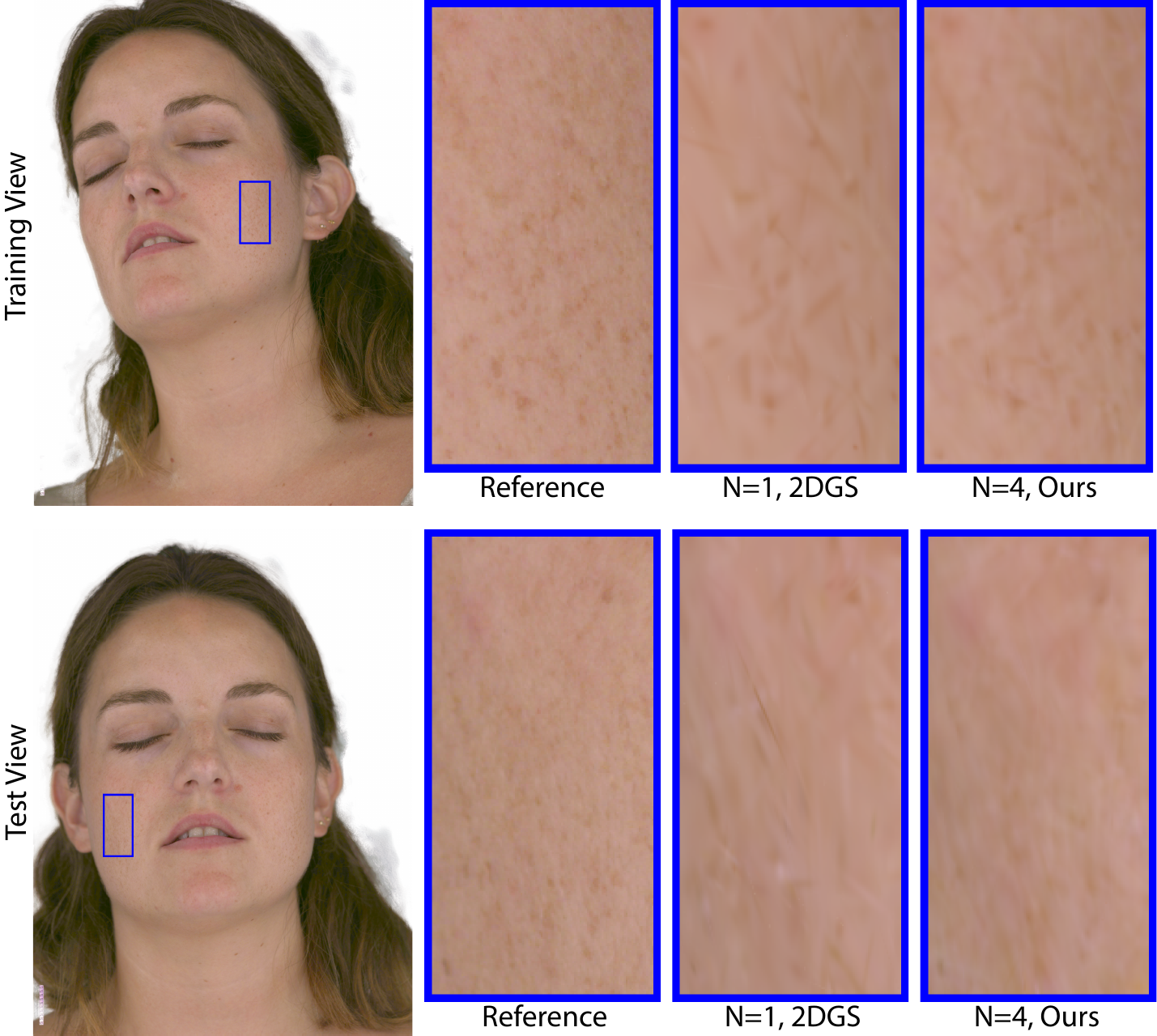}\\%
    {\small
    \begin{tabular}{rr|ccc}\toprule
                               &               & PSNR & SSIM  & LPIPS          \\\midrule
        \multirow{2}{*}{Train} & Baseline 2DGS & 35.9 & 0.928 & 0.159          \\
                               & $N=4$ (ours)  & 35.2 & 0.924 & \textbf{0.151} \\\midrule
        \multirow{2}{*}{Test}  & Baseline 2DGS & 27.6 & 0.866 & 0.273          \\
                               & $N=4$ (ours)  & 27.5 & 0.865 & \textbf{0.263} \\
        \bottomrule
    \end{tabular}}
    \caption{3D Reconstruction of a human face from 14 training cameras and one center test camera. Despite very similar quantitative metrics, our method produces sharper textures on the face, even with less Gaussian primitives, 90566 (Baseline 2DGS) to 64461 (Ours). Images best viewed zoomed-in.}
    \label{fig:medusa}
\end{figure}

The results can be found in \cref{fig:medusa}. Even while quantitative metrics remain almost equal, with only the LPIPS score being noticeably improved from traditional 2DGS ($N=1$) to our method ($N=4$), differences are visible in the qualitative comparison.
Our result preserves more details in both training and testing views while having a more uniform appearance. In the baseline with $N=1$, artifacts produced by elongated splats are especially visible.
Furthermore, with standard densification and pruning enabled, our method achieves better results with fewer primitives ($\approx 90000 \rightarrow {}\approx 64000$) compared to the baseline.
This also reflects in the time to render the test view, which decreases slightly ($0.019s \rightarrow 0.015s$).
The training time, however, increases slightly due to the additional memory transfer in the backward pass (see \cref{sec:method:training}) from 80 minutes to 95 minutes for 30000 epochs.


\section{Conclusion}
\label{sec:conclusion}

In this work we presented an improvement over 2D Gaussian Splatting that replaces the per-primitive solid color with a small color texture that is jointly optimized together with all other 2DGS parameters.
This increases the power to express texture details, leading to an improved image reconstruction for the same number of Gaussian primitives.
Even if the number of primitives can be freely optimized due to densification, splitting, and pruning strategies, our proposed method improves the reconstruction and novel view synthesis over traditional 2DGS on 3D reconstruction tasks.

We envision the proposed approach of \textit{Gaussian Billboards} as an orthogonal approach to many related methods that also tackle the problem of how to enhance the representation quality of 3DGS or 2DGS itself.
In future work, we want to explore the combination of these methods presented recently. Most promisingly, Yu~\etal~\cite{Yu2024-gv} introduce a spatial modulation of the opacity that could show good synergies with our proposed spatial color modulation.

Since the 2DGS primitive attributes are cached in shared memory during the forward and adjoint rasterization stage, the maximal supported color grid size is limited by the available shared memory. In our implementation that extends the CUDA kernels of \textit{gsplat}, this leads to a maximal texture resolution of $8 \times 8$ texels.
In future, we plan to investigate how to lift this technical restriction.

{
    \small
    \bibliographystyle{ieeenat_fullname}
    \bibliography{ms}

\begin{thebibliography}{28}
\providecommand{\natexlab}[1]{#1}
\providecommand{\url}[1]{\texttt{#1}}
\expandafter\ifx\csname urlstyle\endcsname\relax
  \providecommand{\doi}[1]{doi: #1}\else
  \providecommand{\doi}{doi: \begingroup \urlstyle{rm}\Url}\fi

\bibitem[Barron et~al.(2021)Barron, Mildenhall, Verbin, Srinivasan, and
  Hedman]{Barron2021-lg}
Jonathan~T Barron, Ben Mildenhall, Dor Verbin, Pratul~P Srinivasan, and Peter
  Hedman.
\newblock Mip-{NeRF} 360: Unbounded anti-aliased neural radiance fields.
\newblock \emph{arXiv [cs.CV]}, 2021.

\bibitem[Beeler et~al.(2010)Beeler, Bickel, Beardsley, Sumner, and
  Gross]{Beeler2010-jo}
Thabo Beeler, Bernd Bickel, Paul Beardsley, Bob Sumner, and Markus Gross.
\newblock High-quality single-shot capture of facial geometry.
\newblock \emph{ACM Trans. Graph.}, 29\penalty0 (4):\penalty0 1, 2010.

\bibitem[Fang and Wang(2024)]{Fang2024-ug}
Guangchi Fang and Bing Wang.
\newblock Mini-splatting: Representing scenes with a constrained number of
  gaussians.
\newblock \emph{arXiv [cs.CV]}, 2024.

\bibitem[Feng et~al.(2024)Feng, He, Wang, Yang, Kuang, Jun, Fan, and
  Ding]{Feng2024-eq}
Xiang Feng, Yongbo He, Yubo Wang, Yan Yang, Zhenzhong Kuang, Yu Jun, Jianping
  Fan, and Jiajun Ding.
\newblock {SRGS}: Super-resolution {3D} gaussian splatting.
\newblock \emph{arXiv [cs.CV]}, 2024.

\bibitem[Garrido-Jurado et~al.(2014)Garrido-Jurado, Muñoz-Salinas,
  Madrid-Cuevas, and Marín-Jiménez]{Garrido-Jurado2014-pj}
S Garrido-Jurado, R Muñoz-Salinas, F~J Madrid-Cuevas, and M~J Marín-Jiménez.
\newblock Automatic generation and detection of highly reliable fiducial
  markers under occlusion.
\newblock \emph{Pattern Recognit.}, 47\penalty0 (6):\penalty0 2280--2292, 2014.

\bibitem[Huang et~al.(2024)Huang, Yu, Chen, Geiger, and Gao]{Huang2024-2dgs}
Binbin Huang, Zehao Yu, Anpei Chen, Andreas Geiger, and Shenghua Gao.
\newblock 2d gaussian splatting for geometrically accurate radiance fields.
\newblock In \emph{ACM SIGGRAPH 2024 Conference Papers}, New York, NY, USA,
  2024. Association for Computing Machinery.

\bibitem[Höllein et~al.(2024)Höllein, Božič, Zollhöfer, and
  Nießner]{Hollein2024-ca}
Lukas Höllein, Aljaž Božič, Michael Zollhöfer, and Matthias Nießner.
\newblock {3DGS}-{LM}: Faster gaussian-splatting optimization with
  levenberg-marquardt.
\newblock \emph{arXiv [cs.CV]}, 2024.

\bibitem[Jiang et~al.(2023)Jiang, Tu, Liu, Gao, Long, Wang, and
  Ma]{Jiang2023-ax}
Yingwenqi Jiang, Jiadong Tu, Yuan Liu, Xifeng Gao, Xiaoxiao Long, Wenping Wang,
  and Yuexin Ma.
\newblock {GaussianShader}: {3D} gaussian splatting with shading functions for
  reflective surfaces.
\newblock \emph{arXiv [cs.CV]}, 2023.

\bibitem[Kerbl et~al.(2023)Kerbl, Kopanas, Leimkuehler, and
  Drettakis]{Kerbl2023-cv}
Bernhard Kerbl, Georgios Kopanas, Thomas Leimkuehler, and George Drettakis.
\newblock {3D} gaussian splatting for real-time radiance field rendering.
\newblock \emph{ACM Trans. Graph.}, 42\penalty0 (4):\penalty0 1--14, 2023.

\bibitem[Kingma and Ba(2014)]{Kingma2014-fk}
Diederik~P Kingma and Jimmy Ba.
\newblock Adam: A method for stochastic optimization.
\newblock \emph{arXiv [cs.LG]}, 2014.

\bibitem[Li et~al.(2024)Li, Liu, Sznaier, and Camps]{Li2024-nv}
Haolin Li, Jinyang Liu, Mario Sznaier, and Octavia Camps.
\newblock {3D}-{HGS}: {3D} half-gaussian splatting.
\newblock \emph{arXiv [cs.CV]}, 2024.

\bibitem[Lin et~al.(2021)Lin, Ryabtsev, Sengupta, Curless, Seitz, and
  Kemelmacher-Shlizerman]{Lin2021-rn}
Shanchuan Lin, Andrey Ryabtsev, Soumyadip Sengupta, Brian~L Curless, Steven~M
  Seitz, and Ira Kemelmacher-Shlizerman.
\newblock Real-time high-resolution background matting.
\newblock In \emph{Proceedings of the IEEE/CVF Conference on Computer Vision
  and Pattern Recognition}, pages 8762--8771, 2021.

\bibitem[Lombardi et~al.(2021)Lombardi, Simon, Schwartz, Zollhoefer, Sheikh,
  and Saragih]{Lombardi2021-ah}
Stephen Lombardi, Tomas Simon, Gabriel Schwartz, Michael Zollhoefer, Yaser
  Sheikh, and Jason Saragih.
\newblock Mixture of volumetric primitives for efficient neural rendering.
\newblock \emph{ACM Trans. Graph.}, 40\penalty0 (4):\penalty0 1--13, 2021.

\bibitem[Mildenhall et~al.(2021)Mildenhall, Srinivasan, Tancik, Barron,
  Ramamoorthi, and Ng]{Mildenhall2020-mp}
Ben Mildenhall, Pratul~P. Srinivasan, Matthew Tancik, Jonathan~T. Barron, Ravi
  Ramamoorthi, and Ren Ng.
\newblock Nerf: representing scenes as neural radiance fields for view
  synthesis.
\newblock \emph{Commun. ACM}, 65\penalty0 (1):\penalty0 99--106, 2021.

\bibitem[Radl et~al.(2024)Radl, Steiner, Parger, Weinrauch, Kerbl, and
  Steinberger]{Radl2024-mn}
Lukas Radl, Michael Steiner, Mathias Parger, Alexander Weinrauch, Bernhard
  Kerbl, and Markus Steinberger.
\newblock {StopThePop}: Sorted gaussian splatting for view-consistent real-time
  rendering.
\newblock \emph{arXiv [cs.GR]}, 2024.

\bibitem[Rong et~al.(2024)Rong, Chen, Bahmani, Kutulakos, and
  Lindell]{Rong2024-yq}
Victor Rong, Jingxiang Chen, Sherwin Bahmani, Kiriakos~N Kutulakos, and David~B
  Lindell.
\newblock {GStex}: Per-primitive texturing of {2D} gaussian splatting for
  decoupled appearance and geometry modeling.
\newblock \emph{arXiv [cs.CV]}, 2024.

\bibitem[Seiskari et~al.(2024)Seiskari, Ylilammi, Kaatrasalo, Rantalankila,
  Turkulainen, Kannala, Rahtu, and Solin]{Seiskari2024-ss}
Otto Seiskari, Jerry Ylilammi, Valtteri Kaatrasalo, Pekka Rantalankila, Matias
  Turkulainen, Juho Kannala, Esa Rahtu, and Arno Solin.
\newblock Gaussian splatting on the move: Blur and rolling shutter compensation
  for natural camera motion.
\newblock \emph{arXiv [cs.CV]}, 2024.

\bibitem[Wang et~al.(2004)Wang, Bovik, Sheikh, and Simoncelli]{Wang2004-bn}
Zhou Wang, Alan~Conrad Bovik, Hamid~Rahim Sheikh, and Eero~P Simoncelli.
\newblock Image quality assessment: from error visibility to structural
  similarity.
\newblock \emph{IEEE Trans. Image Process.}, 13\penalty0 (4):\penalty0
  600--612, 2004.

\bibitem[Xu et~al.(2024)Xu, Hu, Lai, Shan, and Zhang]{Xu2024-sx}
Tian-Xing Xu, Wenbo Hu, Yu-Kun Lai, Ying Shan, and Song-Hai Zhang.
\newblock Texture-{GS}: Disentangling the geometry and texture for {3D}
  gaussian splatting editing.
\newblock \emph{arXiv [cs.CV]}, 2024.

\bibitem[Yan et~al.(2023)Yan, Low, Chen, and Lee]{Yan2023-ul}
Zhiwen Yan, Weng~Fei Low, Yu Chen, and Gim~Hee Lee.
\newblock Multi-scale {3D} gaussian splatting for anti-aliased rendering.
\newblock \emph{arXiv [cs.CV]}, 2023.

\bibitem[Ye et~al.(2024{\natexlab{a}})Ye, Li, Kerr, Turkulainen, Yi, Pan,
  Seiskari, Ye, Hu, Tancik, and
  Kanazawa]{ye2024gsplatopensourcelibrarygaussian}
Vickie Ye, Ruilong Li, Justin Kerr, Matias Turkulainen, Brent Yi, Zhuoyang Pan,
  Otto Seiskari, Jianbo Ye, Jeffrey Hu, Matthew Tancik, and Angjoo Kanazawa.
\newblock gsplat: An open-source library for {Gaussian} splatting.
\newblock \emph{arXiv preprint arXiv:2409.06765}, 2024{\natexlab{a}}.

\bibitem[Ye et~al.(2024{\natexlab{b}})Ye, Li, Liu, Qiao, and Dou]{Ye2024-db}
Zongxin Ye, Wenyu Li, Sidun Liu, Peng Qiao, and Yong Dou.
\newblock {AbsGS}: Recovering fine details for {3D} gaussian splatting.
\newblock \emph{arXiv [cs.CV]}, 2024{\natexlab{b}}.

\bibitem[Yu et~al.(2024)Yu, Huang, Ling, and Xu]{Yu2024-gv}
Ruihan Yu, Tianyu Huang, Jingwang Ling, and Feng Xu.
\newblock {2DGH}: {2D} gaussian-hermite splatting for high-quality rendering
  and better geometry reconstruction.
\newblock \emph{arXiv [cs.CV]}, 2024.

\bibitem[Yu et~al.(2023)Yu, Chen, Huang, Sattler, and Geiger]{Yu2023-dq}
Zehao Yu, Anpei Chen, Binbin Huang, Torsten Sattler, and Andreas Geiger.
\newblock Mip-splatting: Alias-free {3D} gaussian splatting.
\newblock \emph{arXiv [cs.CV]}, 2023.

\bibitem[Zhang et~al.(2024)Zhang, Zhan, Xu, Lu, and Xing]{Zhang2024-tt}
Jiahui Zhang, Fangneng Zhan, Muyu Xu, Shijian Lu, and Eric Xing.
\newblock {FreGS}: {3D} gaussian splatting with progressive frequency
  regularization.
\newblock \emph{arXiv [cs.CV]}, 2024.

\bibitem[Zhang et~al.(2018)Zhang, Isola, Efros, Shechtman, and
  Wang]{Zhang2018-pz}
Richard Zhang, Phillip Isola, Alexei~A Efros, Eli Shechtman, and Oliver Wang.
\newblock The unreasonable effectiveness of deep features as a perceptual
  metric.
\newblock \emph{CVPR}, 2018.

\bibitem[Zwicker et~al.(2001)Zwicker, Pfister, van Baar, and
  Gross]{Zwicker2001-pc}
M Zwicker, H Pfister, J van Baar, and M Gross.
\newblock {EWA} volume splatting.
\newblock In \emph{Proceedings Visualization, 2001. VIS '01.}, pages 29--538.
  IEEE, 2001.

\bibitem[Zwicker et~al.(2002)Zwicker, Pfister, van Baar, and
  Gross]{Zwicker2002-xh}
M Zwicker, H Pfister, J van Baar, and M Gross.
\newblock {EWA} splatting.
\newblock \emph{IEEE Trans. Vis. Comput. Graph.}, 8\penalty0 (3):\penalty0
  223--238, 2002.

\end{thebibliography}
}

\clearpage
\setcounter{page}{1}
\maketitlesupplementary

\section{Algorithm Details}
In \cref{sec:method:billboards} we introduce the basic algorithm of \textit{Gaussian Billboards}, enhancing 2DGS with spatially-varying textures.
Here we give additional implementation details on how to integrate this into the \textit{gsplat} kernel~\cite{ye2024gsplatopensourcelibrarygaussian}.

In \cref{alg:rasterize_fwd} we show where the bilinear interpolation fits into the rasterization kernels of gsplat with the changes to the original kernel highlighted in \textcolor{algchangedcolor}{green}.
The main changes in the forward algorithm is twofold. First, the parameter fetching is extended from a single color to a grid of color. Second, the bilinear interpolation kernel is invoked per primitive and pixel to obtain the per-pixel color given the current $u,v$.
The central change to the backward algorithm is that the adjoint variables for the $uv$-parameterization, $\hat{u}, \hat{v}$ are now also influenced by the adjoint of the color interpolation, instead of just the Gaussian opacity as in traditional 2DGS.

In \cref{alg:bilinear} we give the detailed pseudocode of the forward and backward implementation of bilinear interpolation.
We manually derive the adjoint code -- the derivatives of the output with respect to all inputs -- from the bilinear interpolation code.
This allows for direct integration of the proposed changes into the fused rasterization kernels of \textit{gsplat}.

\begin{algorithm*}
    \caption{Integration of bilinear color interpolation into the rasterization kernel (Simplified)}
    \label{alg:rasterize_fwd}
    \small
    \begin{algorithmic}[1]
        \Procedure{rasterize\_fwd}{} \Comment{forward code}
        \Statex Each CUDA block processes a range of 2DGS primitives.
        \Statex Each thread per block outputs one pixel.
        \State Compute the pixel location of the current thread $p_x,p_y$
        \State Fetch all primitive parameters of the current range, \textcolor{algchangedcolor}{with the color extended to a color grid}
        \State $T=1.0$ \Comment{The transmittance}
        \State $\mathbf{c}_\text{acc}=(0,0,0)$ \Comment{Accumulated color}
        \ForAll{primitives $k$ in the sorted range}
        \State Intersect the primitive with the view ray $\rightarrow u,v$
        \State \textcolor{algchangedcolor}{Evaluate the color $\mathbf{c}=\text{bilinear}(u,v,N,\sigma,C_k)$}
        \State Compute the opacity $\alpha=\alpha_k*\text{exp}(-\frac{u^2+v^2}{2})$
        \State $\mathbf{c}_\text{acc} \pluseq \mathbf{c} \alpha T$ \Comment{alpha-blending}
        \State $T \asteq (1-\alpha)$
        \EndFor
        \State \textbf{return} $\mathbf{c}_\text{acc}$
        \EndProcedure
        %
        \Statex
        \Procedure{rasterize\_bwd}{} \Comment{backward code}
        \Statex Each CUDA block processes a range of 2DGS primitives.
        \Statex Each thread per block outputs one pixel.
        \State Compute the pixel location of the current thread $p_x,p_y$
        \State Fetch all primitive parameters of the current range, \textcolor{algchangedcolor}{with the color extended to a color grid}
        \State Fetch transmittance $T$ after the last Gaussian
        \State Fetch adjoint of the output color $\hat{\mathbf{c}}_\text{acc}$
        \State $\mathbf{c}_\text{acc}=(0,0,0)$ \Comment{Accumulated color}
        \ForAll{primitives $k$ in the sorted range}
        \State Intersect the primitive with the view ray $\rightarrow u,v$
        \State Compute the opacity $\alpha=\alpha_k*\text{exp}(-\frac{u^2+v^2}{2})$
        \State \textcolor{algchangedcolor}{Evaluate the color $\mathbf{c}=\text{bilinear}(u,v,N,\sigma,C_k)$}
        \Statex ~~~~~~~~~~Adjoint computations:
        \State Adjoint of color with visibility $\hat{\mathbf{c}}_k=(\alpha*T)\hat{\mathbf{c}}_\text{acc}$
        \State \textcolor{algchangedcolor}{Color-adjoint:} $\hat{C},\hat{u},\hat{v}=\text{adj-bilinear}(u,v,C_k,\hat{\mathbf{c}}_k)$
        \State Adjoint of alpha: $\hat{\alpha}$, requires $\mathbf{c}$
        \State Adjoint of Gaussian opacity: $\hat{u},\hat{v}\pluseq ...$
        \State Adjoint of ray intersection: $\hat{\mathbf{p}}, \hat{\mathbf{t}}_u, \hat{\mathbf{t}}_v, \hat{s}_u, \hat{s}_v$
        \State Accumulate gradients into global memory, \textcolor{algchangedcolor}{with the color extended to a color grid}
        \EndFor
        \EndProcedure
    \end{algorithmic}
\end{algorithm*}

\begin{algorithm*}
    \caption{Forward and backward algorithm of bilinear interpolation at $u,v$, given resolution $N$, spatial extend $\sigma$, and an input color grid $C\in\mathbb{R}^{N,N,3}$.}
    \label{alg:bilinear}
    \begin{algorithmic}[1]
        \Function{bilinear}{$u,v, N,\sigma,C$}
        \State $u'=\text{clamp}(\frac{(N-1)*(u+\sigma)}{2*\sigma},0,N-1)$\Comment{From $[-\sigma,+\sigma]$ to $[0,N-1]$}
        \State $v'=\text{clamp}(\frac{(N-1)*(v+\sigma)}{2*\sigma},0,N-1)$
        \State $i_u=\text{floor}(u) \ , \ \ f_u=u'-i_u$ \Comment{Integer index and fractional part }
        \State $i_v=\text{floor}(v) \ , \ \ f_v=v'-i_v$
        \State $\mathbf{c}=$ \Comment{Interpolation}
        \State $\ \ \ C[i_u, i_v, :] * (1-f_u) * (1-f_v) + {}$
        \State $\ \ \ C[i_u+1, i_v, :] * f_u * (1-f_v) + {}$
        \State $\ \ \ C[i_u, i_v+1, :] * (1-f_u) * f_v + {}$
        \State $\ \ \ C[i_u+1, i_v+1, :] * f_u * f_v$
        \State \textbf{return} $\mathbf{c}$
        \EndFunction

        \Function{adj-bilinear}{$u,v, N,\sigma,C, \hat{\mathbf{c}}$}
        \Statex ~~~$\hat{c}$ indicates the adjoint variable of variable $c$
        \State $u'=\text{clamp}(\frac{(N-1)*(u+\sigma)}{2*\sigma},0,N-1)$\Comment{From $[-\sigma,+\sigma]$ to $[0,N-1]$}
        \State $v'=\text{clamp}(\frac{(N-1)*(v+\sigma)}{2*\sigma},0,N-1)$
        \State $i_u=\text{floor}(u) \ , \ \ f_u=u'-i_u$ \Comment{Integer index and fractional part }
        \State $i_v=\text{floor}(v) \ , \ \ f_v=v'-i_v$
        \Statex ~~~Adjoint for color:
        \State $\hat{C}[i_u, i_v] = \hat{\mathbf{c}} * (1-f_u) * (1-f_v)$
        \State $\hat{C}[i_u+1, i_v] = \hat{\mathbf{c}} * f_u * (1-f_v)$
        \State $\hat{C}[i_u, i_v+1] = \hat{\mathbf{c}} * (1-f_u) * f_v$
        \State $\hat{C}[i_u+1, i_v+1] = \hat{\mathbf{c}} * f_u * f_v$
        \Statex ~~~Adjoint for fractional grid position $u,v$ (uses dot-product):
        \State $\hat{f}_u = \hat{\mathbf{c}} \ \bigcdot \ \left( (C[i_u+1, i_v, :]-C[i_u, i_v, :]) * (1-f_v) + (C[i_u+1, i_v+1, :]-C[i_u, i_v+1, :]) * f_v\right)$
        \State $\hat{f}_v = \hat{\mathbf{c}} \ \bigcdot \ \left( (C[i_u, i_v+1, :]-C[i_u, i_v, :]) * (1-f_u) + (C[i_u+1, i_v+1, :]-C[i_u+1, i_v, :]) * f_u\right)$
        \Statex ~~~Adjoint for $u,v$ coordinates:
        \State $\hat{u} = (u>-\sigma \text{ and } u<+\sigma) \ ? \ \left( \frac{(N-1)}{2\sigma} \hat{f}_u \right) \ : \ 0$
        \State $\hat{v} = (v>-\sigma \text{ and } v<+\sigma) \ ? \ \left( \frac{(N-1)}{2\sigma} \hat{f}_v \right) \ : \ 0$
        \State \textbf{return} $\hat{C}, \hat{u}, \hat{v}$
        \EndFunction
    \end{algorithmic}
\end{algorithm*}


\end{document}